%% file: main.tex
\documentclass[10pt,journal,compsoc]{IEEEtran}

\usepackage[nocompress]{cite}
\usepackage{tabularx}
\usepackage{colortbl}
\usepackage{graphicx}
\usepackage{booktabs}
\usepackage[table]{xcolor}
\usepackage{colortbl}
\usepackage{amsmath,amsfonts,dsfont}
\usepackage{amsmath}

\usepackage{algorithmic}
\usepackage{algorithm}
\usepackage{array}
\usepackage{textcomp}
\usepackage{stfloats}
\usepackage{multirow}
\usepackage{url}
\usepackage{verbatim}
\usepackage{graphicx}
\usepackage{xcolor}
\usepackage{units}

\usepackage{subcaption}   

\usepackage [numbers,sort,square]{natbib}

\makeatletter

\usepackage{hyperref}  
\usepackage{booktabs}
\usepackage{stfloats}
\usepackage{adjustbox}
\usepackage{colortbl}
\definecolor{F7E0D5}{RGB}{247,224,180}
\colorlet{Light}{white!0!F7E0D5}

\definecolor{cb-green}     {RGB}{   0,  128,  073}

\ifCLASSINFOpdf
\else
\fi
\begin{document}
\title{A Unified Masked Jigsaw Puzzle Framework for Vision and Language Models}

\author{Weixin~Ye,
        Wei~Wang,
        Yahui~Liu,
        Yue~Song,
        Bin~Ren,
        Wei~Bi,
        Rita~Cucchiara,
        and~Nicu~Sebe,~\IEEEmembership{Senior~Member,~IEEE}
\IEEEcompsocitemizethanks{\IEEEcompsocthanksitem Weixin Ye and Wei Wang are with Beijing Jiaotong University, Beijing, China. Yahui Liu is with Kuaishou, Beijing, China. Wei Bi is with Hong Kong University of Science and Technology, Hong Kong, China. Yue Song is with Caltech, USA. Nicu Sebe and Bin Ren are with the University of Trento, Italy. Rita Cucchiara is with University of Modena and Reggio Emilia, Modena, Italy. Wei Wang is the corresponding author.\protect\\
E-mail: \{weixinye, wei.wang\}@bjtu.edu.cn, yahui.cvrs@gmail.com, \{yue.song, nicu.sebe, bin.ren\}@unitn.it, weibi@ust.hk, rita.cucchiara@unimore.it}
\thanks{Manuscript received October 05, 2025; revised March 27, 2025.}}


\markboth{Journal of \LaTeX\ Class Files,~Vol.~14, No.~8, August~2015}%
{Shell \MakeLowercase{\textit{et al.}}: Bare Advanced Demo of IEEEtran.cls for IEEE Computer Society Journals}
\input{sections/0abstract}

\maketitle
\IEEEdisplaynontitleabstractindextext
\IEEEpeerreviewmaketitle

\input{sections/1introduction}
\input{sections/2relatedworks}
\input{sections/3framework}
\input{sections/4experiments}

\input{sections/5conclusion}
\ifCLASSOPTIONcaptionsoff
  \newpage
\fi


%

%

\bibliography{ref}

\vspace{-1.3cm}
\begin{IEEEbiography}[{\includegraphics[width=1in,height=1.25in,clip,keepaspectratio]{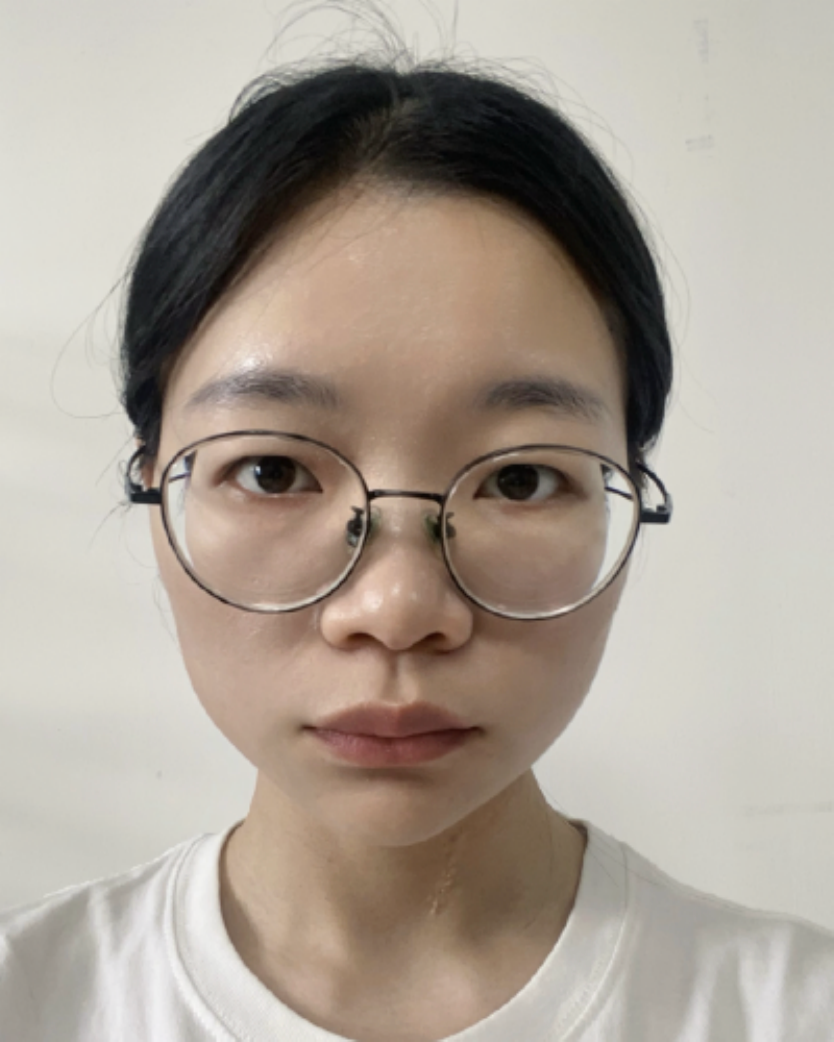}}]{Weixin Ye} is a third-year Ph.D student with the School of Computer Science and Technology, Beijing Jiaotong University. She received the M.Eng degree in Shihezi university in 2023. Her research interests are computer vision, deep learning, and natural language processing.
\end{IEEEbiography}
\vspace{-1.3cm}
\begin{IEEEbiography}[{\includegraphics[width=1in,height=1.25in,clip,keepaspectratio]{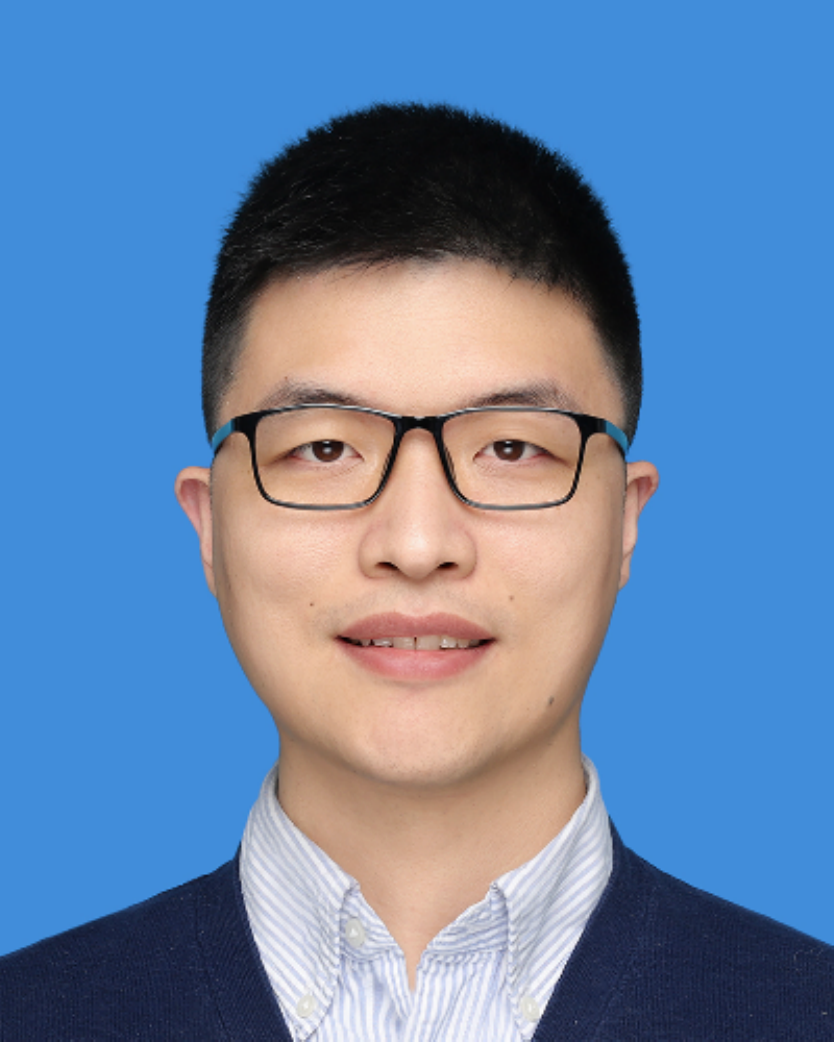}}]{Wei Wang}
is a Professor of Computer Science at University of Beijing Jiaotong University. Previously, after obtaining his PhD from University of Trento in 2018, supervised by Prof. Nicu Sebe in Multimedia and Human Understanding Group (MHUG), he became a Postdoc at EPFL, Switzerland. His research interests include machine learning and its application to computer vision and multimedia analysis.
\end{IEEEbiography}
\vspace{-1.3cm}
\begin{IEEEbiography}[{\includegraphics[width=1in,height=1.25in,clip,keepaspectratio]{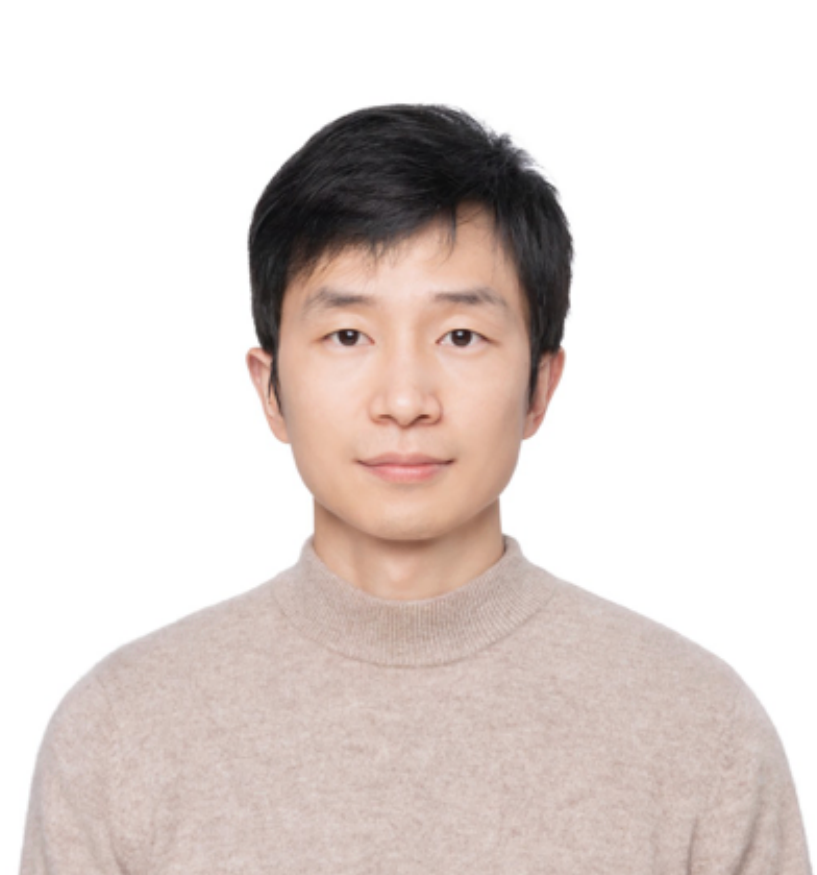}}]{Yahui Liu} is now a Principal Engineer at Kuaishou, Beijing, China. He got his certificate of Doctor Degree from the Multimedia and Human Understanding Group (MHUG) at the University of Trento, Italy, supervised by Prof. Nicu Sebe and Dr. Bruno Lepri. He received B.Eng. degree and M.Eng. degree from Wuhan University. His research interests lie in the areas of Computer Vision and Natural Language Processing. 
\end{IEEEbiography}
\vspace{-1.3cm}
\begin{IEEEbiography}[{\includegraphics[width=1in,height=1.25in,clip,keepaspectratio]{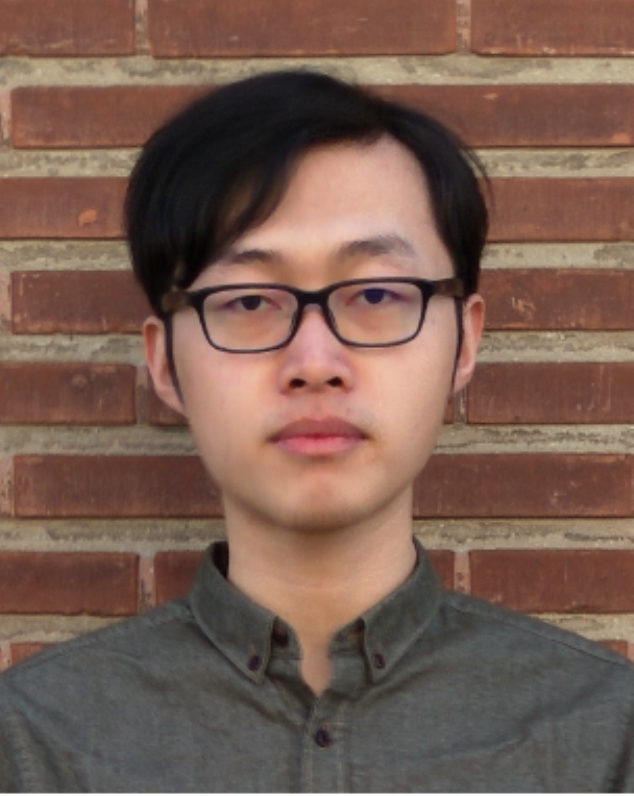}}]{Yue Song} received the B.Sc. \emph{cum laude} from KU Leuven, Belgium and the joint M.Sc. \emph{summa cum laude} from the University of Trento, Italy and KTH Royal Institute of Technology, Sweden, and the Ph.D. \emph{summa cum laude} from the Multimedia and Human Understanding Group (MHUG) at the University of Trento, Italy. Currently, he is a post-doctoral research associate at Caltech. His research interests are structured representation learning.
\end{IEEEbiography}
\vspace{-1.3cm}
\begin{IEEEbiography}[{\includegraphics[width=1in,height=1.25in,clip,keepaspectratio]{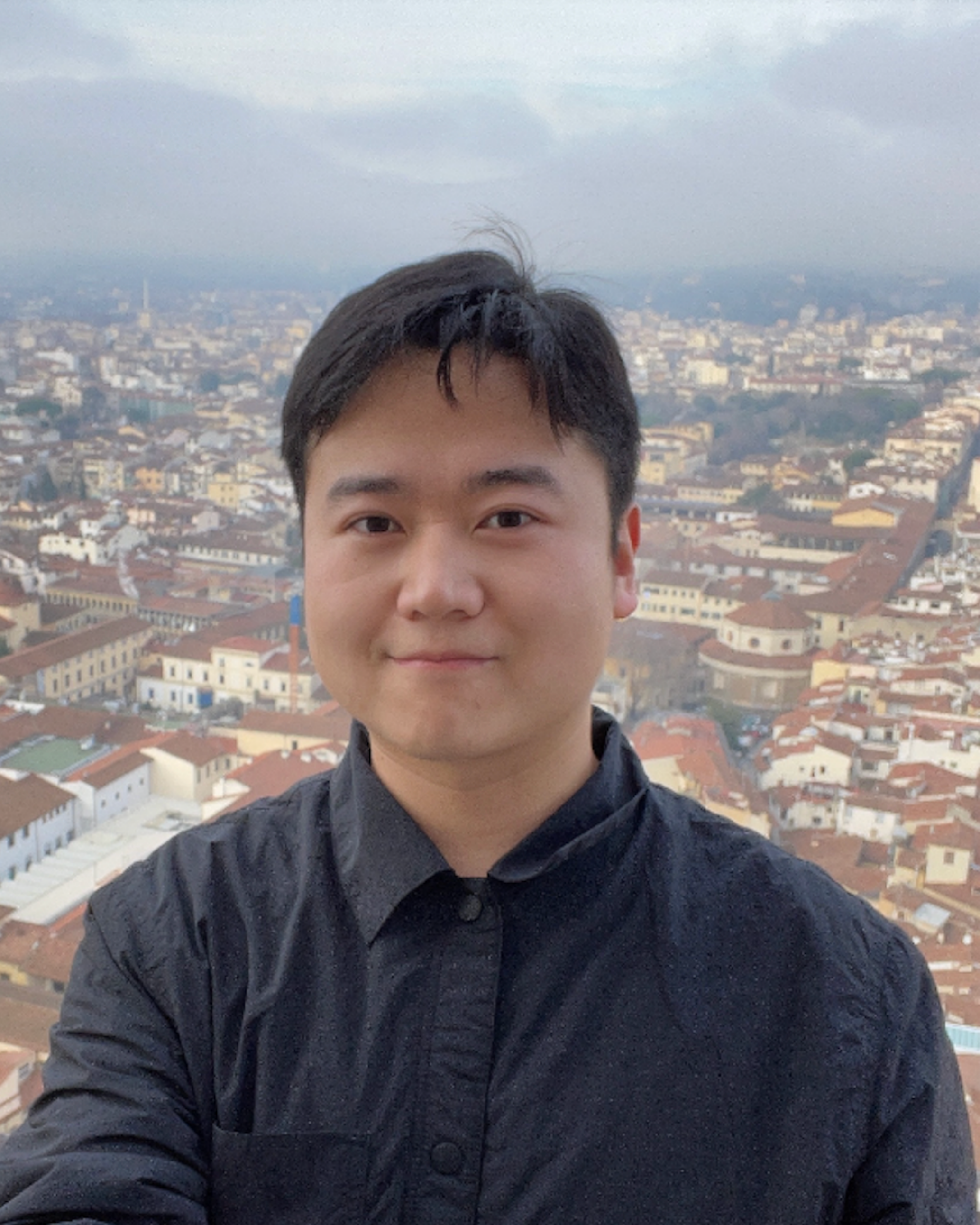}}]{Bin Ren}
is a second-year Ph.D. student in the Italian National Artificial Intelligence Program co-organized by the University of Pisa and the University of Trento. He is in the Multimedia and Human Understanding Group (MHUG) advised by Prof. Dr. Nicu Sebe and Prof. Dr. Rita Cucchiara. His research interests lie in the intersection of computer vision and deep learning.
\end{IEEEbiography}
\vspace{-1.3cm}
\begin{IEEEbiography}[{\includegraphics[width=1in,height=1.25in,clip,keepaspectratio]{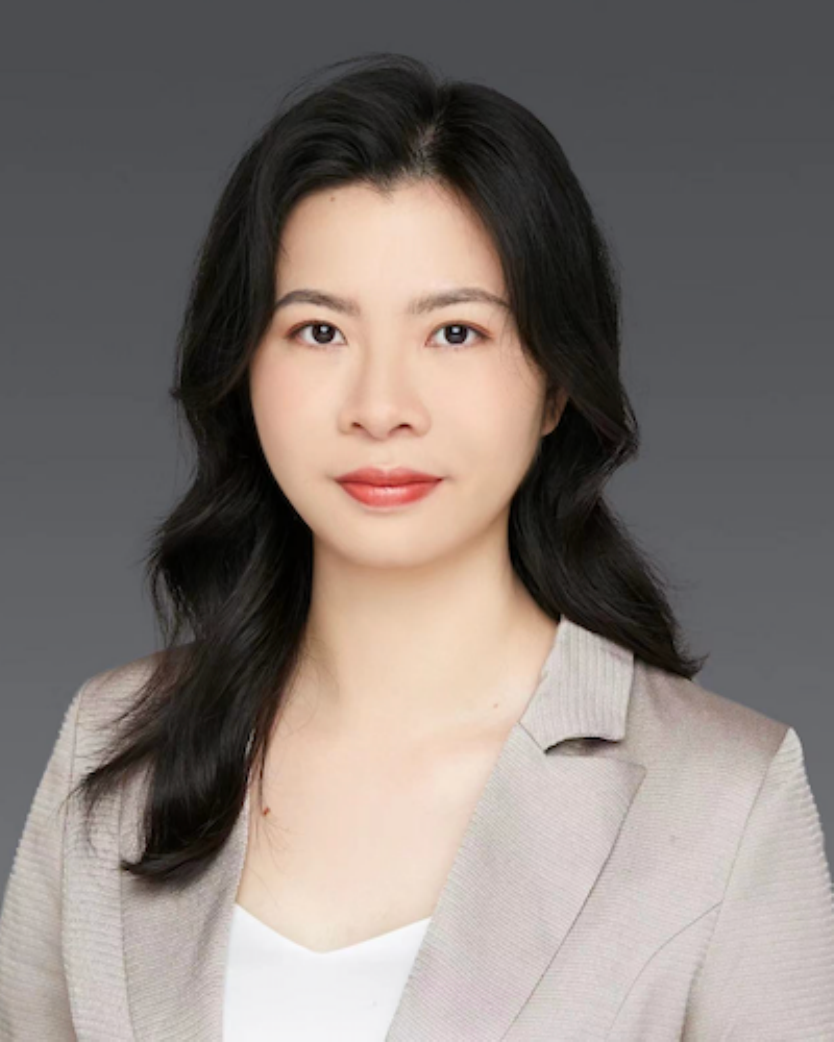}}]{Wei Bi}is a researcher at Hong Kong University of Science and Technology. She received her Ph.D. in Computer Science and Engineering from the Hong Kong University of Science and Technology in 2015. She is an awardee of the Google Ph.D. Fellowship in 2013 and the Google Anita Borg Scholarship in 2014. Her research interests lie in the broad areas of machine learning, natural language processing, and artificial intelligence. 
\end{IEEEbiography}
\vspace{-1.3cm}
\begin{IEEEbiography}[{\includegraphics[width=1in,height=1.25in,clip,keepaspectratio]{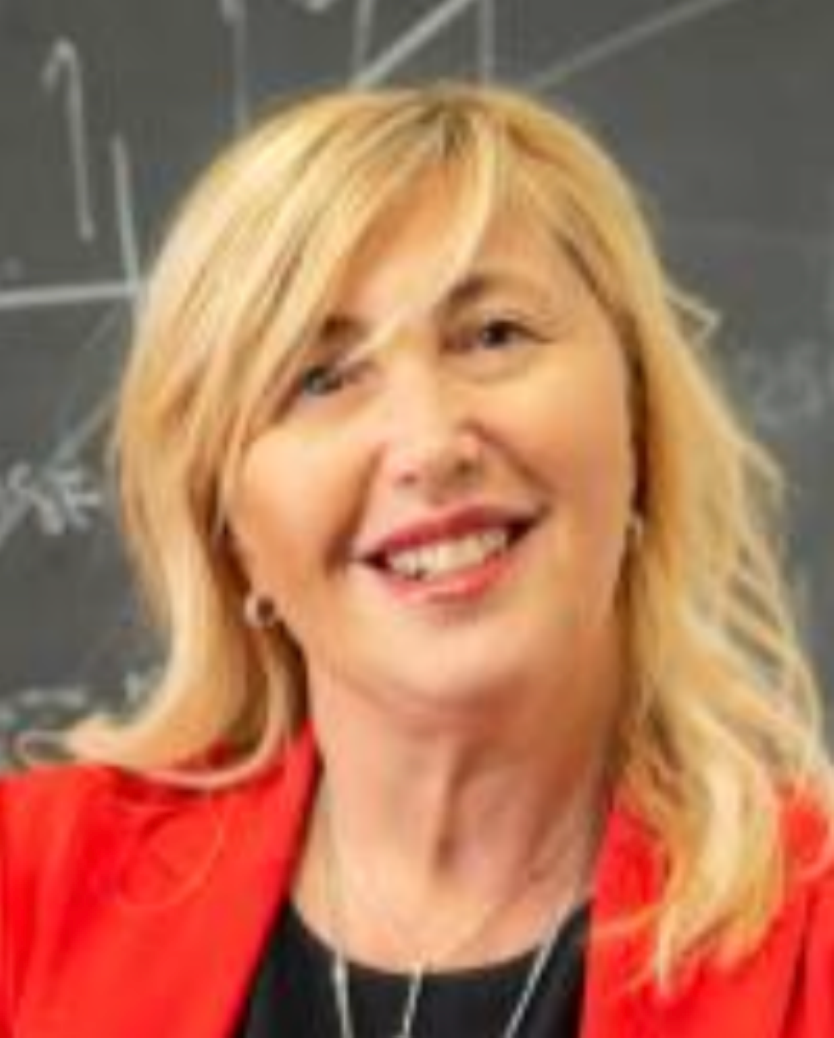}}]{Rita Cucchiara} received the M.Sc. degree in Electronics Engineering and the Ph.D. degree in Computer Engineering from the University of Bologna. She is a Full Professor at the University of Modena and Reggio Emilia, leading the AImage-Lab. She has authored or coauthored over 400 papers in journals and international proceedings. She is Member of the Advisory Board of the Computer Vision Foundation, and Director of the ELLIS Unit of Modena.


\end{IEEEbiography}
\vspace{-1.3cm}
\begin{IEEEbiography}[{\includegraphics[width=1in,height=1.25in,clip,keepaspectratio]{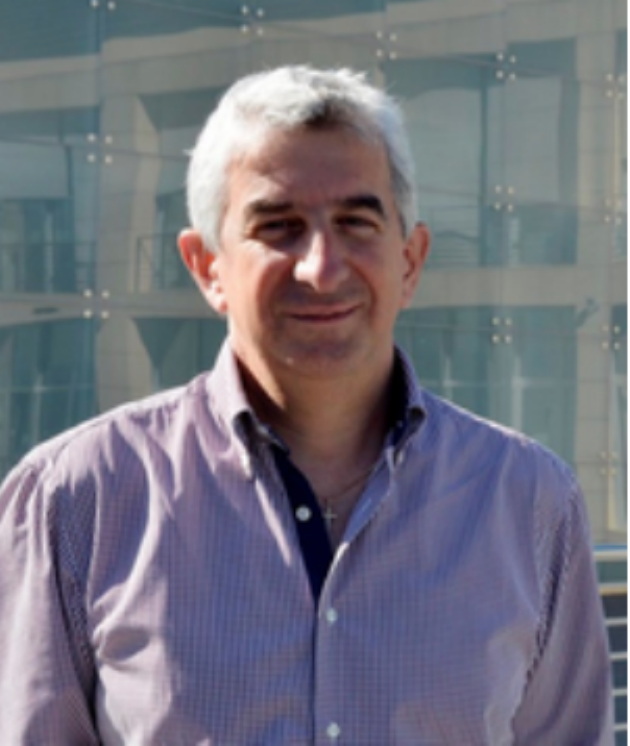}}]{Nicu Sebe} (Senior Member, IEEE) is a Professor at the University of Trento, Italy, leading the research in the areas of multimedia information retrieval and human behavior understanding in Multimedia and Human Understanding Group (MHUG). He was the General Co- Chair of ACM Multimedia 2013, and the Program Chair of ACM Multimedia 2007 and 2011, ECCV 2016, ICCV 2017 and ICPR 2020. He is a fellow of the International Association for Pattern Recognition.
\end{IEEEbiography}







\end{document}

%% file: sections/0abstract.tex
\IEEEtitleabstractindextext{

\begin{abstract}
In federated learning, Transformer, as a popular architecture, faces critical challenges in defending against gradient attacks and improving model performance in both Computer Vision (CV) and Natural Language Processing (NLP) tasks. It has been revealed that the gradient of Position Embeddings (PEs) in Transformer contains sufficient information, which can be used to reconstruct the input data. To mitigate this issue, we introduce a Masked Jigsaw Puzzle (MJP) framework. MJP starts with random token shuffling to break the token order, and then a learnable \textit{unknown (unk)} position embedding is used to mask out the PEs of the shuffled tokens. In this manner, the local spatial information which is encoded in the position embeddings is disrupted, and the models are forced to learn feature representations that are less reliant on the local spatial information. Notably, with the careful use of MJP, we can not only improve models' robustness against gradient attacks, but also boost their performance in both vision and text application scenarios, such as classification for images (\textit{e.g.,} ImageNet-1K) and sentiment analysis for text (\textit{e.g.,} Yelp and Amazon).
Experimental results suggest that MJP is a unified framework for different Transformer-based models in both vision and language tasks. 
\textcolor{black}{Code is publicly available via ~\url{https://github.com/ywxsuperstar/transformerattack}}
\end{abstract}

}


%% file: sections/1introduction.tex
\ifCLASSOPTIONcompsoc
\IEEEraisesectionheading{\section{Introduction}\label{sec:introduction}}
\else
\section{Introduction}
\label{sec:introduction}
\fi

\begin{figure}[!h]
    \centering
    \includegraphics[width=0.85\linewidth]{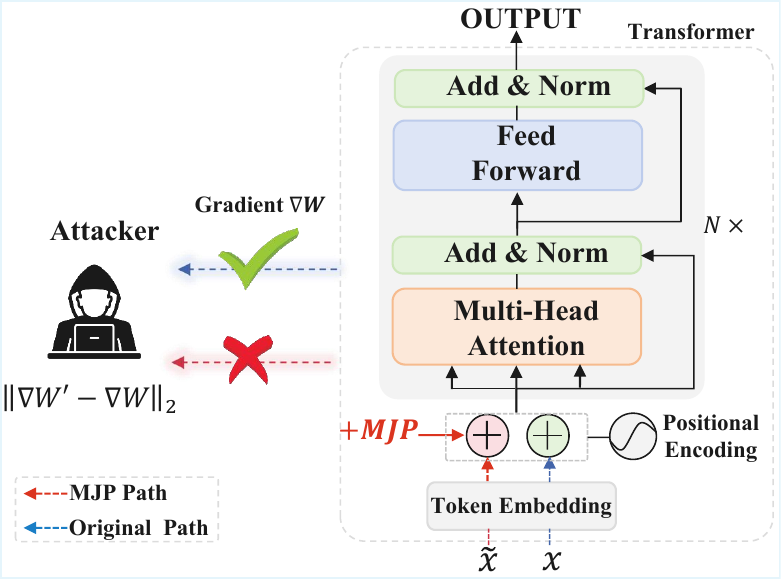}
    \vspace{-2mm}
    \caption{\textcolor{black}{Illustration of privacy leakage through Transformer model gradients. An attacker reconstructs input data by minimizing the distance between the real gradients $\nabla{\mathbf{W}}$ and the dummy gradients $\nabla{\mathbf{W}'}$. This leakage is mitigated by shuffling the input data and employing the MJP method to disrupt the spatial information of the PEs while preserving model performance, with $x$ and $\tilde{x}$ denoting the original and shuffled data, respectively.}} 
    \label{fig:privacy_attacker} 
    \vspace{-5mm}
\end{figure}

\IEEEPARstart{F}{ederated} learning (FL) has emerged as a powerful paradigm for training machine learning models across distributed devices, which allows multiple clients to collaboratively train models without sharing their private data to provide an extra level of user data privacy protection~\cite{10180041,Brisimi_Chen_Mela_Olshevsky_Paschalidis_Shi_2018}. However, this decentralized nature also exposes several vulnerabilities, particularly related to privacy. One of the potential threats in federated learning is gradient inversion attacks, where adversaries could exploit shared gradients to attempt the reconstruction of sensitive user data, such as images or text, as illustrated in Fig. ~\ref{fig:privacy_attacker}. This has led to growing concerns about privacy in federated learning, especially when Transformer-based models are involved, which are widely used in both Computer Vision (CV) and Natural Language Processing (NLP).

\begin{figure}[!h]
    \centering
    \includegraphics[width=1.0\linewidth]
    {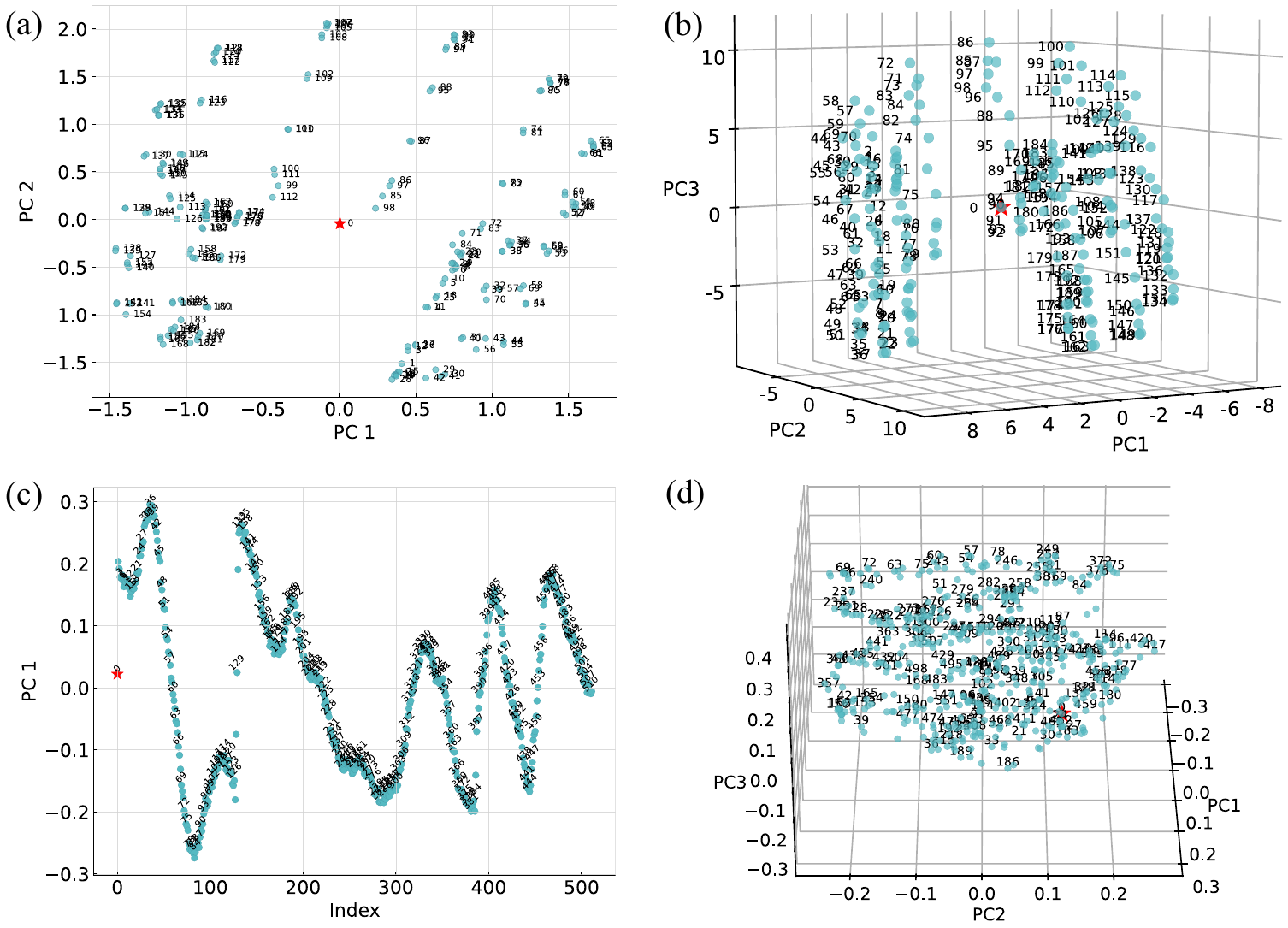} 
    \vspace{-5mm}
    \caption{Low-dimensional projection of PEs. (a) The 2D PCA projection (DeiT-S~\cite{touvron2021training}), it shows that the PEs from neighboring positions tend to cluster together, which reveals a consistent pattern that reflects the spatial relationships of the input patch positions. (b) The 3D PCA projection (DeiT-S~\cite{touvron2021training}), it also shows that the position information is well captured with PEs.(c) The 1D PCA projection (BERT$_\text{BASE}$~\cite{2018BERT}), a similar sinusoidal pattern shows the same order as the input token positions. (d) The 3D PCA projection (BERT$_\text{BASE}$~\cite{2018BERT}), position information can be captured by PEs. Note that the embedding of index 0 (highlighted in red) corresponds to the first token embedding.}
    \label{fig:origin_posemb}
    \vspace{-5mm}
\end{figure}

\begin{figure*}[!h]
    \centering
    \includegraphics[width=7in]{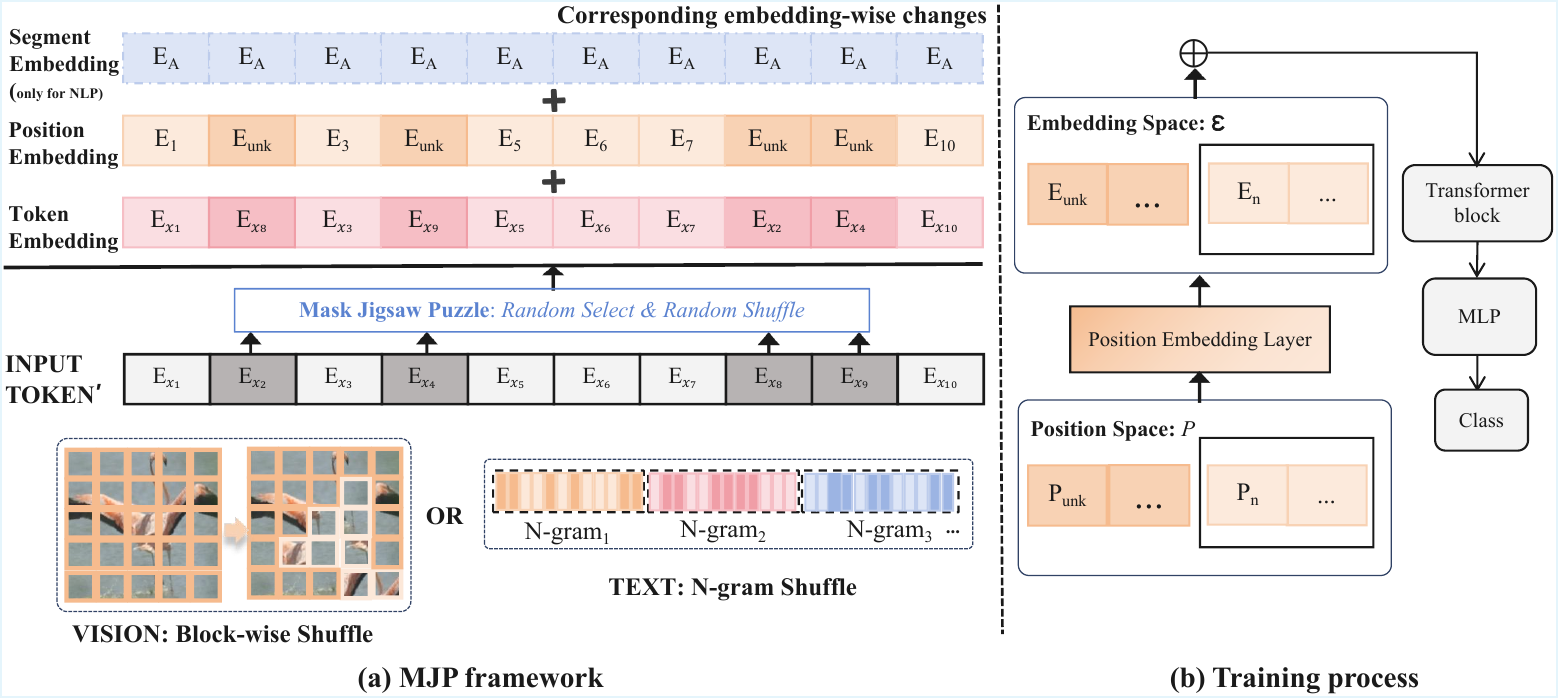}
    \vspace{-2mm}
    \caption{(a) A unified MJP framework: We feed the randomly shuffled tokens (vision for block-wise shuffling and text for \textit{n}-gram shuffling) into the model, and the PEs of these selected tokens are masked correspondingly. Note that segment embedding is only used in NLP tasks. Dark yellow indicates selected PEs are masked with \textit{unknown} (unk), and the rest of PEs remain unchanged. (b) Training process: After MJP operation, the embeddings are fed to the model directly. Note that the \textit{unk} embeddings are defined as parameters which are learned in the training process.}
    \label{fig:overview-all}  
    \vspace{-5mm}
\end{figure*}

The Transformer~\cite{vaswani2017attention} has become remarkably effective in deep learning, and has achieved significant advancements in both CV and NLP tasks. In the field of CV, advances in Transformers have driven progress in image classification~\cite{dosovitskiy2020image,touvron2021training}, object detection~\cite{carion2020end,zhu2021deformable}, segmentation~\cite{ye2019cross} and image generation~\cite{chen2021pre,liang2021swinir}. In the field of NLP, Transformers have demonstrated exceptional performance across a range of tasks~\cite{2018BERT}, including text classification~\cite{yang2020xlnet}, named entity recognition~\cite{yang2020xlnet}, machine translation~\cite{raffel2023exploring}, and question answering~\cite{brown2020language,khan2022transformers}. 
Naturally, this success has led to a surge of research focused on improving Transformer architectures, especially the self-attention mechanisms (SAMs)~\cite{liu2021swin,wang2022pvt,wang2021pyramid} for Vision Transformer (ViT) models~\cite{fang2021you,gidaris2018unsupervised,girdhar2019video} and pre-trained language models (PLMs)~\cite{2018BERT,raffel2023exploring,Radford_Narasimhan_Salimans_Sutskever}.
In contrast, Position Embeddings (PEs) receive less attention from the research community and have not been well studied yet. In fact, these embeddings, which encode the positional or spatial relationships of input tokens, are crucial for the proper functioning of Transformers. It has been demonstrated that, in the absence of PEs, the pure language Transformer encoders (\emph{e.g.}, BERT~\cite{devlin2018bert} and RoBERTa~\cite{liu2019roberta}) may not well capture the positional structure information~\cite{wang2020position}. As a consequence, the information of a sentence cannot be well represented~\cite{dufter2022position}. Similarly, in the vision domain, Dosovitskiy \emph{et al.}~\cite{dosovitskiy2020image} reveal that removing PEs causes performance degradation. \textcolor{black}{However, recent studies have revealed a significant privacy concern: the spatial information encoded in PEs may inadvertently expose sensitive data. As demonstrated by Lu \emph{et al.}\cite{lu2022april} and Ren \emph{et al.}\cite{ren2023masked}, these PEs allow attackers to reconstruct original input data by exploiting the shared gradients in FL. The vulnerability stems from the fact that the complete spatial relationships are preserved, as shown \textit{Original Path} in Fig. ~\ref{fig:privacy_attacker}. This presents a serious privacy risk, especially in decentralized FL settings where user data should remain private.}



\textcolor{black}{To explore the impacts of PEs, it is essential to discern explicitly what positional information PEs extract from input tokens.} 
For vision, we visually demonstrate that the PEs of DeiT-S~\cite{touvron2021training} can learn the 2D and 3D spatial relationship very well from the input image patches. The relationship is visualized in Fig.~\ref{fig:origin_posemb} (a) and (b). For text, we project the high-dimensional PEs of BERT$_\text{BASE}$~\cite{2018BERT} into 1D and 3D spaces using Principal Component Analysis (PCA), as visualized in Fig.~\ref{fig:origin_posemb} (c) and (d). In both vision and language models, we observe that the PEs are distributed in the same organized order as the input token positions. This alignment explains why PEs bring performance gains for Transformer-based models ~\cite{dosovitskiy2020image, 2018BERT}, as the spatial relationships work
similarly to the inherent inductive bias found in Convolutional Neural Networks (CNNs) (\textit{i.e.,} local visual structures)~\cite{xu2021vitae}. Unfortunately, these well-learned spatial relationships are also the key factors resulting in privacy leakage~\cite{lu2022april}. Based on these observations, a straightforward idea to protect user privacy is to provide transformers with randomly transformed (\emph{i.e.}, shuffled) input data. The intuition is that the original correct spatial relationships within input tokens will be violated via such a transformation. Therefore, we transform the previous visually recognizable input tokens $\mathbf{x}$ into their unrecognizable counterparts $\widetilde{\mathbf{x}}$ during the training. As demonstrated in the bottom left of Fig.~\ref{fig:overview-all}(a), it is easy to recognize that the patch tokens (on the left) are parts of a crane. After shuffling, it becomes difficult to tell the content of the patches (on the right). 
As such, introducing incorrect spatial information into the original data can mislead potential attacks, making it more difficult to recover the original input data, as shown shuffled input data $\tilde{x}$ in Fig. ~\ref{fig:privacy_attacker}. Experimental results indicate that this strategy can effectively alleviate the privacy leakage. However, this naive strategy leads to severe accuracy drop and loss of spatial information. 
To address these issues, we need to incorporate operations that can preserve the essential features of the data, maintaining model performance while simultaneously reducing privacy leakage.

In this paper, we propose a unified Masked Jigsaw Puzzle (MJP) framework for Transformer-based models, with an overview shown in Fig.~\ref{fig:overview-all}. 
Our approach begins with a token-wise masking technique that randomly selects a portion of the input sequential tokens. Then, we apply a jigsaw puzzle operation to these selected tokens where their order is shuffled. Next, we mask the PEs of these shuffled tokens by a shared unknown position embedding. We employ the MJP method with the introduction of Dense Absolute Location (DAL) loss to enhance the relationship among unshuffled tokens for image classification task on ImageNet-1K. For text tasks, we enhance the MJP framework by introducing an \textit{n}-gram window, which restricts tokens shuffling within individual windows. This approach disrupts the local spatial information without harming global features, which consistently shows performance improvements over baseline models such as Transformer4 and BERT\textsubscript{BASE} on Yelp and Amazon. Empirically, we demonstrate that the proposed framework MJP can balance the trade-off between model performance and user privacy in Transformer-based models. \textcolor{black}{The \textit{MJP Path} in Fig.~\ref{fig:privacy_attacker} outlines our privacy-preserving pipeline, which effectively mitigates gradient leakage while maintaining model utility.} MJP's applications and further details in both image and text tasks are provided in  Sec.~\ref{sec:framework}. We further have a thorough discussion to reveal insights of how MJP works across different domains and demonstrate its effectiveness in privacy protect, and extensive ablation studies.

The preliminary result of this work is published at~\cite{ren2023masked}.
In the conference paper, our proposed Masked Jigsaw Puzzle (MJP) position embedding showed that for vision tasks, it can simultaneously improve accuracy, consistency, and user privacy protection. This journal extension proposes that these techniques can also be applied to text tasks, which are termed as a unified MJP framework. Compared to the conference version, we include the following new contents: 1) apart from vision tasks, we have also successfully applied MJP approach to language models for sentiment analysis; 2) we have implemented optimization-based image attacks and demonstrate that MJP can effectively mitigate gradient leakage with DAL loss; 3) MJP can effectively defend against gradient inversion attacks for language models on long texts (\emph{i.e.}, Yelp and Amazon) by introducing a \textit{n}-gram window; 4) the proposed unified MJP framework can not only protect user privacy but also boost performance for both vision and language models.

The rest of the paper is organized as follows: \textcolor{black}{Sec.~\ref{sec:related-work} describes the PEs in Transformers and their applications in vision, text, and privacy protection.}
Sec.~\ref{sec:framework} shows specific implementation details of the MJP framework. 
Sec.~\ref{sec:experiments} conduct extensive experiments in vision and text tasks which explains why the MJP framework can improve accuracy and alleviate user privacy leakage. We also provide detailed experimental results and some in-depth analysis. Finally, Sec.~\ref{sec:discussion and conclusion} discuss the work and summarize the conclusions.


%% file: sections/2relatedworks.tex
\section{Related Work}
\label{sec:related-work}
\noindent\textbf{Transformer Model.}
Benefiting from the strong representation power of modeling global relationships in both vision and text domains, Vision Transformers (ViTs)~\cite{dosovitskiy2020image} and Pre-trained Language Models (PLMs) (\emph{e.g.}, BERT\cite{2018BERT}) have achieved superior performance than their counterpart CNNs/RNNs on various downstream tasks. In the domain of vision, Transformers have been applied to object detection~\cite{carion2020end,zhu2021deformable}, object re-identification~\cite{he2021transreid}, dense prediction~\cite{yang2021transformer,zheng2021rethinking,wang2021end,wang2022tokencut}, image generation~\cite{chen2020generative,chen2021pre,liang2021swinir,jiang2021transgan}), \emph{etc}. For language tasks, Transformers have been applied to sentiment analysis\cite{2013Recursive}, named entity recognition\cite{2016Neural}, and question answering\cite{rajpurkar2016squad}, \emph{etc}.

\noindent\textbf{Position Embeddings.}
In Transformers, both the attention and (individual token-based) feed-forward layers are permutation invariant when the position information is not considered. To address this limitation, PEs are introduced to provide information about the order of tokens, which enables the capture of dependencies between elements at different positions. Unlike RNNs, Transformers inherently lack awareness of sequence order. Therefore, it is crucial to add PEs to distinguish identical words in different positions in both vision and text tasks. The work~\cite{vaswani2017attention} introduced both fixed and trainable PEs, which enhanced model performance in tasks such as machine translation and text generation. Subsequent research, including innovations like relative positional encoding~\cite{shaw2018selfattention} and applications in BERT, has further refined these techniques, underscoring the critical role of PEs in Transformer-based models. Previous works~\cite{gehring2017convolutional,vaswani2017attention,shaw2018self} have indicated that PEs are useful in providing the model with a sense of which portion of the input or output sequence it is currently processing. Inspired by this, some works~\cite{jung2020global,su2021roformer,kiyono2021shape,liu2022petr,he2022masked} showed diverse application scenarios that benefit from the usage of suitable PEs. In addition, Chu \emph{et al.}~\cite{chu2021conditional} proposed correlating PEs with their local neighborhood of the input sequence. Liu \emph{et al.}~\cite{liu2021efficient} proposed to enhance the spatial prior (\emph{i.e.}, relative localization) in the final content embedding to indirectly enrich the inductive bias. Obviously, although these methods enhance the position information learned by PEs, they indeed degenerate the position-insensitive property of multi-head self-attentions. In this paper, we dive into the usage of PEs and extend MJP to improve the position-insensitive property of Transformers without hurting the positive effects of PEs.

\noindent\textbf{Spatial Priors in PEs.}
To visualize the concrete relationships of input tokens captured in the high-dimensional PEs, we use PCA to project them into a lower-dimensional space. 
When we reduce the dimensionality of PEs from models (BERT$_\text{BASE}$~\cite{2018BERT} and DeiT-S~\cite{touvron2021training}), the features of PEs are rotated and projected onto new axes (principal components, PCs), which capture the most variance in the data. 
\textcolor{black}{The coordinate axes in Fig. ~\ref{fig:origin_posemb} are labeled with the PCs, where PC1 (first PC) captures the highest variance in the data, PC2 (second PC) the second-highest, and so on.
Fig. ~\ref{fig:origin_posemb} (a) and (b) show the 2D and 3D PCA projections for DeiT-S~\cite{touvron2021training}, where neighboring positions tend to cluster together, indicating that the model generates similar embeddings for adjacent positions. 
Fig. ~\ref{fig:origin_posemb} (c) and (d) display projections of PEs for BERT$_\text{BASE}$~\cite{devlin2018bert}. In the 1D PCA projection, the embeddings form a sinusoidal curve, while in 3D, they adopt a helical shape, with adjacent positions clustering similarly, which reflects the relationships among tokens.
These observations confirm that the spatial relationships in the high-dimensional space are preserved in the lower-dimensional space. The learnable position embeddings, which act as a lookup table mapping 1D data into sparse high-dimensional space, form a large and sparse matrix. Dimensionality reduction effectively captures the structural information of these sparse matrices even in lower dimensions. \textit{For the visualization of relationships in PE, we annotate the plots with the corresponding position indices for each projected PE. For DeiT-S, we label every position. For BERT$_\text{BASE}$ (with a maximum length of 512), we label every third position to ensure readability.}}


\noindent\textbf{Gradient Inversion Attack.}
User privacy is a paramount concern in federated learning due to the involvement of sensitive information from individual clients in the collaborative training process. Gradient leakage poses a significant privacy threat by exploiting the shared gradient to reconstruct the original input data. Previous research has demonstrated that the gradient can be leaked in both vision~\cite{Zhu2019dlg,zhao2020idlg, yin2021gradients,hatamizadeh2022gradvit,lu2022april,geiping2020invertinggradientseasy} and text~\cite{Zhu2019dlg, deng2021tag} tasks, which reveals the vulnerability of CNN and Transformer models to gradient-based inversion attacks. To mitigate gradient leakage, traditional methods, such as differential privacy~\cite{wei2019federatedlearningdifferentialprivacy} and secure multi-party computation~\cite{goldwasser1997multi} have been introduced; however, these often involve trade-offs in terms of model performance and computational efficiency. Recently, Gao \textit{et al.}~\cite{Gao2020PrivacypreservingCL} employed data augmentation to obscure sensitive information, while Soteria \textit{et al.}~\cite{sun2020provabledefenseprivacyleakage} pruned gradients in a single layer. Both approaches directly manipulate the gradients to preserve privacy while maintaining model performance. Different from these methods, our MJP method disturbs the spatial information of PEs to defend against the gradient inversion attack.

%% file: sections/3framework.tex
\section{Framework}
\label{sec:framework}
As a fundamental component in Transformer architectures, multi-head self-attentions (MSAs)~\cite{vaswani2017attention,dosovitskiy2020image} aggregate sequential tokens with normalized attentions, which are formulated as $\pmb{z}_j = \sum_i \texttt{Softmax}(\frac{\pmb{Q}\pmb{K}}{\sqrt{d}})_i\pmb{V}_{i,j}$, where $\pmb{Q}$, $\pmb{K}$ and $\pmb{V}$ are query, key and value matrices, respectively. $d$ is the dimension of the query and key, and $\pmb{z}_j$ is the $j$-th output token. Theoretically, when the position information is not considered, the outputs of MSAs should be invariant to the order of input sequence (\emph{i.e.}, \emph{position-insensitivity or position-agnostic}). 
This implies that the original input data can be recognized by the Transformer-based model, even if the data is transformed into an unrecognizable form by permuting the order of tokens. Intuitively, when a model extracts similar representations from both the original data and its transformed versions, the single gradient can lead to multiple potential predictions, which complicates the process of the gradient inversion.

\begin{algorithm}[t]
	\renewcommand{\algorithmicrequire}{\textbf{Input:}}
	\renewcommand{\algorithmicensure}{\textbf{Output:}}
	\caption{Token-wise Random Jigsaw Puzzle Shuffle}
	\label{alg:1}
	\begin{algorithmic}[1]
		\REQUIRE Input sequence: $\mathbf{x}\in \mathbb{R}^{N \times L}$; \\  ~~~~~~Shuffle Ratio: $\gamma$; 
		~~~~~~Token Size: $L$
		\ENSURE Shuffled tokens: $\widetilde{\pmb{x}}_t$
		\STATE 
		~~~~~~$\mathbf{x}_t\in\mathbb{R}^{N\times L} \longleftarrow 
        Split(\mathbf{x})
  $
		\STATE ~~~~~~$\mathbf{m}\in\mathbb{R}^{N\times L} \longleftarrow BinaryInitialize(\mathbf{x}_t, 0)$
		\STATE 
		~~~~~~$\widetilde{\mathbf{m}} \in\mathbb{R}^{N\times L} \longleftarrow TokenwiseMask(\mathbf{m}, \mathbf{\gamma})$
	    \STATE
	    ~~~~~~$\widetilde{\pmb{x}}_t \in\mathbb{R}^{N\times L} \longleftarrow JigsawPuzzle(\mathbf{x}_t, \widetilde{\mathbf{m}})$
		\STATE 
		\textbf{return} $\widetilde{\pmb{x}}_t$
	\end{algorithmic}  
\end{algorithm}
\vspace{-4mm}
However, the usage of PEs hinders such implementations. The PEs inject positional information into the tokens such that the outputs of the transformer vary dramatically with the mentioned naive transformations (\emph{i.e.} breaking the token order by shuffling). 
To this end, we propose a token-wise random jigsaw puzzle shuffle algorithm (See Alg.~\ref{alg:1}) to transform the input tokens with different shuffle ratios $\gamma$ for intermingling the original correct spatial relationship. 
In this section, we focus on demonstrating the MJP unified framework and its application in both vision and text tasks.


\subsection{Masked Jigsaw Puzzle Shuffle}
\label{sec:token_shuffle} 
Given the sequential input $\mathbf{x}\in \mathbb{R}^{N \times L}$, where $L$ is the length of the original input sequence and $N$ represents the batch size. $L_{max}$ deontes the maximal length of the input sequence padded or truncated for batch processing. We first split the sequence into tokens $\mathbf{x}_t$ (\textit{i.e.} tokenize image or sentence), and then we initialize a binary mask matrix $\mathbf{m}\in\mathbb{R}^{N \times L_{max} }$ with the same size as the sequence $\mathbf{x}$. Next, we use \textit{Token-wise Mask} to update the binary mask $\mathbf{m}$, in which the masked positions will be set to 1 and the rest untouched positions remain 0. \textit{Token-wise Mask} (line 3 of Alg.~\ref{alg:1}) is defined as the positions of the selected shuffled tokens that need to be masked. The hyper-parameter $\gamma$ is used to control the ratio of selected tokens. After that, a jigsaw puzzle shuffle operation is applied to $\mathbf{x}_t$ conditioned on the updated binary mask $\widetilde{\mathbf{m}}$ (line 4 of Alg.~\ref{alg:1}). In the \textit{JigsawPuzzle} operation, we randomly permute orders of the tokens with the $\{\widetilde{\mathbf{m}}_{n}^{l} = 1 | n\in\{1,2,\dots,N\}, l\in\{1,2,\dots,L\}\}$, where $n$ refers to the $n$-th sequence in the batch data and $l$ refers to the $l$-th token in this sequence. Finally, we get the shuffled token sequence $\widetilde{\mathbf{x}}_t$.  

Notably, in vision applications, our \textit{Token-wise Mask} draws inspiration from the original Block-wise Masking approach~\cite{bao2021beit}. Whereas the original method randomly masks the sequence tokens, and then these masked tokens are not visible to the encoder module. On the contrary, our masking strategy only masks out the corresponding positions of the selected tokens. These selected tokens are still visible to the encoder module but are randomly shuffled in a jigsaw puzzle manner. The details of the jigsaw puzzle shuffling are presented in Sec~\ref{subsec: mjp on vision and text}.

\subsection{MJP Position Embedding}
\label{subsec:mjp}

Typically, the input sequential tokens are projected to token/patch embeddings, by the projection operations (\textit{i.e.,} a tokenizer and an embedding layer for language models, and a linear mapping layer for vision models). Given the original input tokens $\mathbf{x}$ in the Transformer-based model, we can formulate the function of the input layer, which is located before the Transformer block, as follows:
\begin{equation}
\mathbf{z}_{0} = [\mathbf{x}_t^1\mathbf{E}; \mathbf{x}_t^2\mathbf{E}, \cdots, \mathbf{x}_t^L\mathbf{E}] + \mathbf{E}_{\text{pos}},
\end{equation}
where $\mathbf{E}$ represents the projection layer; $\mathbf{E}_{\text{pos}}{\in}\mathbb{R}^{L\times D}$ denotes PEs; $D$ means the dimension of the patch/token embedding. Note that we usually increase $L$ by 1 here for additional [CLS] embedding in vision tasks.

Next, we apply the proposed \textit{Token-wise Mask} algorithm to $\mathbf{x}_t$ produce the transformed token sequences $\widetilde{\mathbf{x}}_t$ with the shuffle ratio $\gamma$. In this scenario, if we retain the original position embedding sequence, it leads to a mismatch issue between the shuffled tokens and the PE sequence. To address this issue, we introduce a shared \textit{unknown} position embedding for the shuffled positions to alleviate the mismatching issue. With the corresponding updated mask $\widetilde{\mathbf{m}}$ from Alg.~\ref{alg:1}, we propose our MJP PEs as follows:
\begin{equation}
\widetilde{\mathbf{E}}_{\mathrm{pos}}^l= \begin{cases}\mathbf{E}_{\mathrm{pos}}^l, & \text { if } \widetilde{\mathbf{m}}_n^l=0 \\ \mathbf{E}_{\mathrm{unk}}, & \text { if } \widetilde{\mathbf{m}}_n^l=1\end{cases},
\end{equation}
where $\mathbf{E}_{\text{unk}}{\in}\mathbb{R}^{1\times D}$ denotes shared \textit{unknown} position embedding. It represents that the sequence token in this position has random permutation and its position should be occluded (line 3 of Alg.~\ref{alg:1}).
$\mathbf{E}_{\text{pos}}$ is the original position embedding for the remaining sequence tokens. Thus, we update the input layer with a new formulation:
\begin{equation}
\label{Eq:input_layer}
\widetilde{\mathbf{z}}_{0} = [\widetilde{\mathbf{x}}_t^1\mathbf{E}; \widetilde{\mathbf{x}}_t^2\mathbf{E}, \cdots, \widetilde{\mathbf{x}}_t^L\mathbf{E}] + \widetilde{\mathbf{E}}_{\text{pos}},
\end{equation} 
the following procedures and modules are exactly the same as the ones in the original transformer. A toy illustration of the proposed MJP is shown in Fig.~\ref{fig:overview-all}(a), where dark yellow and dark red represent the randomly permuted tokens and \emph{unknown} PEs, while light yellow and light red indicate the rest of regular tokens and its original PEs.

\subsection{Applications in Vision and Text}
\label{subsec: mjp on vision and text}
\subsubsection{Vision Models}
\label{subsec:dal}
Liu \emph{et al.}~\cite{liu2021efficient} noticed that by enhancing the 2-D spatial information of the output embeddings in the last layer of ViTs, the training convergence speed can be accelerated. Inspired by their work, we observe that similar gains can be obtained by applying a low-dimensional spatial prior in PEs. Different from the dense relative localization constraint in~\cite{liu2021efficient}, which samples relative pairs from the whole output sequential embeddings, we propose a simple yet efficient \emph{dense absolute localization} (DAL) regression method. This method directly employs self-supervised absolute location to enhance the spatial information in PEs, and does not need to sample relative pairs as in~\cite{liu2021efficient}.  

For vision tasks, since the PEs capture the absolute position of the input patches, to some extent, the position information could be reconstructed via a reversed mapping function $g(\cdot):\mathcal{E} \to \mathcal{P}$, where $\mathcal{E}$ and $\mathcal{P}$ are embedding space and position space, respectively. Given that PEs have one-to-one correspondence with the sequential image patches, we can reshape them into $\mathbf{E}_{\text{pos}}\in\mathbb{R}^{K\times K \times D}$, where $K$ refers to height/width of the grid ($K^2=L$), and $D$ refers to latent vector size. Then we can compute the reverse mapping from $\mathcal{E} \to \mathcal{P}$ via:
\begin{equation} 
(\widetilde{i},\widetilde{j})^T = g(\mathbf{E}^{i,j}_{\text{pos}}),
\end{equation}
in which $(\widetilde{i},\widetilde{j})^T$ denotes the predicted patch position, and $\mathbf{E}^{i,j}_{\text{pos}}$ is the position embedding of the patch $(i,j)$ in the $K\times K$ grid.
The dense absolute localization (DAL) loss is defined as:
\begin{equation}
\label{Eq:dal}
\mathcal{L}_{\text{DAL}} = \mathbb{E}_{\mathbf{E}^{i,j}_{\text{pos}}, 1\leq i,j \leq K}[\|(i,j)^T - (\widetilde{i},\widetilde{j})^T\|_1],
\end{equation}
where the expectation is computed by averaging the $\ell_1$ loss between the correspond $(i, j)^T$ and $(\widetilde{i},\widetilde{j})^T$. Then, $\mathcal{L}_{\text{DAL}}$ is added to the standard cross-entropy loss ($\mathcal{L}_{\text{CE}}$) of the naive ViTs. The final loss is: $\mathcal{L}_{\text{all}} = \mathcal{L}_{\text{CE}} + \lambda \mathcal{L}_{\text{DAL}}$, where we set $\lambda =$ 0.01 for all vision experiments.
We also formalize these procedures as an algorithm in Alg.~\ref{alg:2}. 

\begin{algorithm}[!h]
	\renewcommand{\algorithmicrequire}{\textbf{Input:}}
	\renewcommand{\algorithmicensure}{\textbf{Output:}}
	\caption{The pipeline of the proposed MJP for vision.}
	\label{alg:2}
	\begin{algorithmic}[1]
		\REQUIRE Input sequence: $\mathbf{x}\in \mathbb{R}^{L}$; \\ 
        ~~~~~~Shuffle Ratio: $\gamma$; 
        ~~~~~~Token Size: $L$ 
		\STATE 
		$\widetilde{\pmb{x}}_t \leftarrow Alg. ~\ref{alg:1}(\mathbf{x}, L, \gamma)$ 
		\STATE 
		$\mathbf{E}_{\mathrm{unk}}(\widetilde{\pmb{x}}_t)$ 
  \IF{$training$}
		\STATE
		$\mathbf{DAL}(\mathbf{x} - \mathbf{x} \cap        
  \widetilde{\pmb{x}}_t)$ 
    \ENDIF
	\end{algorithmic}  
\end{algorithm}
\vspace{-2mm}

\subsubsection{Text Models}
\label{subsec:text-models}
For text models, inspired by the original \textit{n}-gram window concept~\cite{1992Class}, which slices a string into a set of overlapping \textit{n}-grams\cite{cavnar1994n}, we adapt the \textit{Token-wise Mask} method for text tasks by introducing an \textit{n}-gram window during shuffling tokens. 
Given a sequential sentence, we first divide the original tokens into \textit{n}-grams with the same size (except for the last one) by a sliding window without overlap, and then apply the \textit{Token-wise Mask} within each window independently. For simplification, 
we define $\mathbf{S}$ to represent the operations in $Alg. ~\ref{alg:1}$. 
Then, we can re-formulate the proposed MJP method in Eq.~(\ref{Eq:input_layer}) for the \textit{n}-gram-based shuffling tokens in input layer as:
\begin{equation}
\small
\widetilde{\mathbf{z}}_{0} = \left[ \sum_{i=0}^{\frac{L}{w}-1} \mathbf{S}( \mathbf{x}_t^{iw+1:(i+1)w}, \gamma, w)\mathbf{E} \right]+ \widetilde{\mathbf{E}}_{\text{pos}},
\end{equation}
where $w$ denotes the size of the \textit{n}-gram window, and $i$ represents index of the sliding window. An intuitive example is shown the left part of Fig.~\ref{fig:overview-all} (\textit{N-gram shuffle} process), where the dark color tokens represent the shuffled tokens, while the light color tokens remain unchanged. Actually, the working principle of \textit{n}-gram windows is similar to a Hamming window, which disturbs the order of local tokens resulting in smaller changes of spatial information. Simultaneously, the order of each \textit{n}-gram window is unchanged, which preserves global coherence to improve model performance. Compared to shuffling the whole sentence, the division of \textit{n}-gram windows only disrupts the local spatial information while preserving the global features. 

Additionally, considering that token encoding is based on absolute positions, we explore the DAL regressor loss function. Similar to Eq.~\ref{Eq:dal}, we optimize the distance between predicted and original token positions by introducing DAL to reinforce the relationships between unshuffled tokens. Referring to the work~\cite{liu2021efficient}, we also adapt DAL into Dense relative location regressor (DRL) loss, which employs only the relative position to predict the location of tokens, and other processes remain the same as DAL. In this situation, the final loss is: $\mathcal{L}_{\text{all}} = \mathcal{L}_{\text{CE}} + \lambda \mathcal{L}_{\text{aux}}$, where we set $\lambda =$ 0.01, and $\mathcal{L}_{\text{aux}}$
is either $\mathcal{L}_{\text{DAL}}$ or $\mathcal{L}_{\text{DRL}}$.

%% file: sections/4experiments.tex
\section{Experiments}
\label{sec:experiments}
\subsection{Experimental Setup}
\subsubsection{Overview}
\label{sec:overview_setup}
\textcolor{black}{
To empirically validate the effectiveness of the MJP mechanism, we adopt a two-stage experimental procedure (the (\textbf{non-federated} learning \& \textbf{federated} learning setups). }

\noindent\textbf{\textcolor{black}{Non-federated learning setup:}}
\textcolor{black}{In this stage, the MJP is applied to the baseline model evaluated under a centralized (\textbf{non-federated}) learning setup. Performance is evaluated by comparing the results with and without the MJP integrationacross multiple downstream tasks, including image classification, image segmentation, textual sentiment analysis, and multiple-choice question answering. }

\noindent\textbf{\textcolor{black}{Federated learning setup:}} \textcolor{black}{In this stage, MJP is incorporated into a \textbf{federated} learning framework~\cite{lu2022april,deng2021tag} to evaluate its privacy-preserving capabilities. We simulate a privacy attack where an adversary attempts to reconstruct input data from the shared gradients. To align with the first stage, the victim client in this gradient inversion attack conducts classification tasks.The effectiveness of the attack and the defense is then quantified by evaluating the semantic content of the data reconstructed by the adversary, using the detailed metrics outlined in Section~\ref{sec:setup_metric}. This stage directly assesses the capability of MJP to mitigate privacy leakage under a realistic threat model.}

\subsubsection{Image Modality}
We follow the configurations from the previous work~\cite{ren2023masked} to validate the proposed MJP framework. All experiments, including image classification and image attack tasks, are conducted on the ImageNet-1K~\cite{russakovsky2015imagenet} dataset. 
In these experiments, we utilize the famous ViTs (\emph{e.g.}, Swin~\cite{liu2021swin}, DeiT~\cite{touvron2021training}), for image classification in a standard supervised learning manner. For the ImageNet-1K experiments, we employ the AdamW~\cite{kingma2015adam} optimizer for 300 epochs using a cosine decay learning rate scheduler and 20 epochs for linear warm-up. The training is conducted with a batch size of 1024, an initial learning rate of 0.001, and a weight decay of 0.05. \textcolor{black}{We also incorporate various augmentation and regularization strategies (e.g., repeated augmentation~\cite{hoffer2020augment}, CutMix~\cite{yun2019cutmix}, and Mixup~\cite{zhang2018mixup}) during the training}.
\subsubsection{Text Modality}
\noindent\textbf{Datasets.} 
For the experiments on text tasks, we verify the proposed MJP framework on Yelp~\cite{zhang2016characterlevel} and Amazon~\cite{hou2024bridging} datasets. The Yelp is a five-class sentiment classification dataset, and Amazon dataset is about \textit{Movies\_and\_TV} reviews that contain five stars level. Given the scarcity of very long and short sentences, we select sentences with lengths between 180 and 400 words for both datasets to capture a central portion of the data's distribution, ensuring statistical representativeness and avoiding outliers. We constructed a subset of the Yelp dataset consisting of 166,365 sentences and split it with a training-to-validation ratio of approximately 9:1 (\emph{i.e.}, 149,728 sentences for training and 16,637 sentences for validation). Similarly, we select a subset of Amazon including 553,512 sentences in total and use the same split ratio (\emph{i.e.}, 498,160 sentences for training and 55,352 sentences for validation). 

\noindent\textbf{Training setup.}
For text tasks, we evaluate our approach on a simple Transformer-based model with four stacked Transformer layers (Transformer4)~\cite{deng2021tag} and BERT$_\text{BASE}$~\cite{devlin2018bert}. We employ the pre-trained BERT-Base-Uncased~\cite{DBLP:journals/corr/abs-1810-04805} model for fine-tuning BERT$_\text{BASE}$. \textcolor{black}{The learning rate is set to $5\times10^{-3}$ for the classification layer, and $5\times10^{-5}$ for other layers. The batch size is set to 16. For weight decay, we apply $1\times10^{-3}$ to all parameters except bias and LayerNorm parameters (\texttt{bias}, \texttt{LayerNorm.bias}, and \texttt{LayerNorm.weight})}. We use the AdamW optimizer~\cite{kingma2015adam} for 4 epochs with linear warm-up, where the number of warm-up steps is set to 10\% of the total steps. Gradient clipping is applied with a maximum norm of 1 during training to prevent exploding gradients. The window size is set to 32 for all models. For Transformer4, we train from scratch with a learning rate of $5\times10^{-3}$ for 8 epochs.

\subsubsection{Evaluation Metrics}
\label{sec:setup_metric}
\noindent\textbf{Model Performance with MJP.} As stated in the introduction section, our goal is to enhance model performance or at least have less negative impact on the model performances, whether during the finetuning process or training from scratch. For all experiments, we use the widely used Top-1 accuracy (\textbf{Top-1 Acc.}) to indicate whether the model performance is invariant to the jigsaw puzzle transformation.

\noindent\textbf{Gradient Inversion with MJP.} For image recovery, we introduce image similarity metrics that account for pixel-wise mismatch to evaluate the anti-attack performance of a model, including Mean Square Error (MSE), Peak Signal-to-Noise Ratio (PSNR), cosine similarity in the Fourier space (FFT$_\text{2D}$), and Learned Perceptual Image Patch Similarity (LPIPS)~\cite{zhang2018unreasonable}. Higher values of MSE, FFT$_\text{2D}$, and LPIPS, along with lower values of PSNR and SSIM indicate greater discrepancies between the original and recovered images. For text reconstruction, we introduce sentence similarity metrics to evaluate sentence similarity, containing Rouge1, Rouge2, RougeL~\cite{lin2004rouge}, G-bleu\cite{papineni-etal-2002-bleu}, Sacrebleu (S-bleu) ~\cite{2002BLEU} and token accuracy (Token$_{\text{ac}}$)~\cite{deng2021tag} for computing semantic similarity. The Token$_{\text{ac}}$ measures the overlap between the reconstructed and the true sentence~\cite{deng2021tag}. Metrics such as Rouge and BLEU are higher when the reconstructed text closely aligns with the original.

\begin{figure*}[!ht]
    \centering
    \begin{subfigure}[b]{0.3\textwidth}
        \centering
        \includegraphics[width=0.9\linewidth]{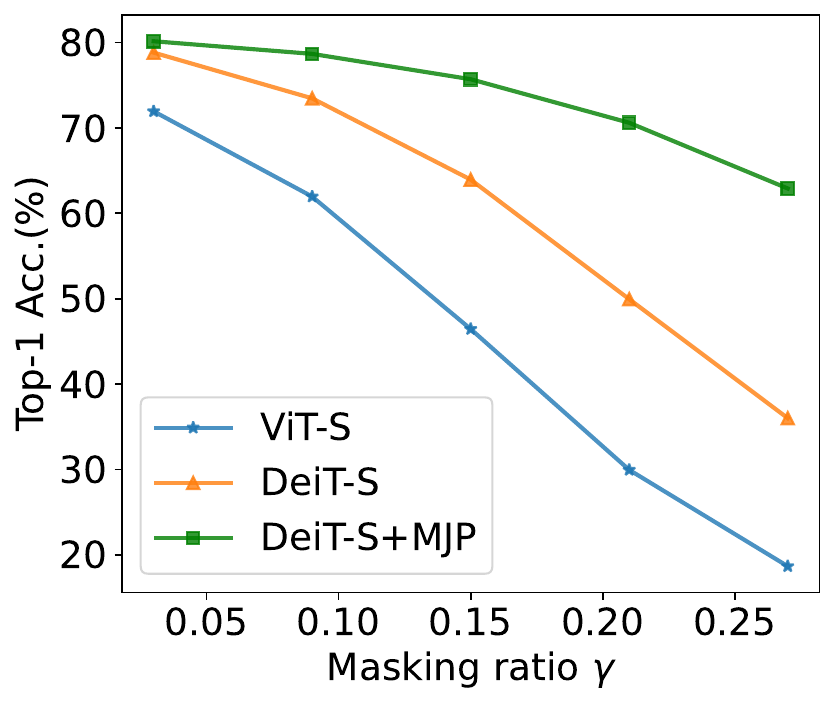}
        \vspace{-3mm}
        \caption{}
    \end{subfigure}
    \hfill
    \begin{subfigure}[b]{0.3\textwidth}
        \centering
        \includegraphics[width=0.9\linewidth]{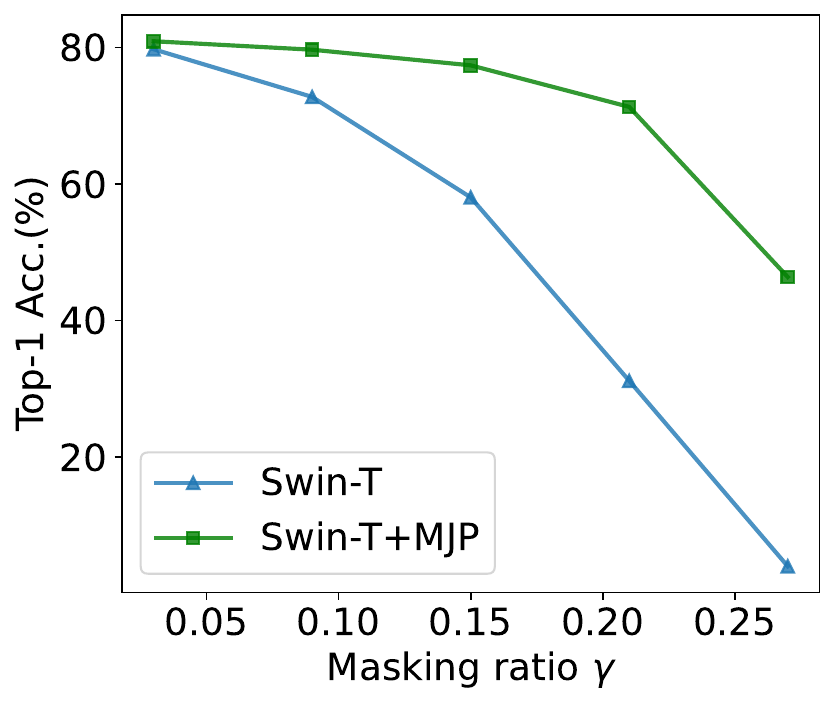}
        \vspace{-3mm}
        \caption{}
    \end{subfigure}
    \hfill
    \begin{subfigure}[b]{0.3\textwidth}
        \centering
        \includegraphics[width=0.9\linewidth]{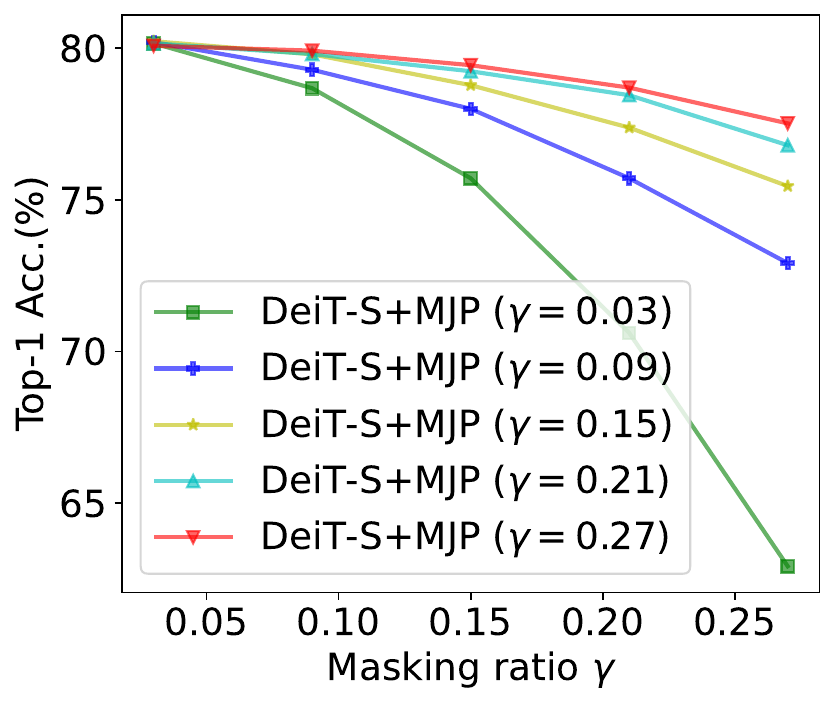}
        \vspace{-3mm}
        \caption{}
    \end{subfigure}
    \vspace{-3mm}
    \caption{Ablation on the mask ratio $\gamma$ during inference: (a) comparisons among ViT-S, DeiT-S, and our method (trained with $\gamma=0.03$); (b) comparisons between Swin-T and our method (trained with $\gamma=0.03$); (c) comparisons of our method on DeiT-S trained with different $\gamma$.}
    \label{fig:masking-ratio}
    \vspace{-5mm}
\end{figure*}

Note that in typical gradient inversion methods, the goal is to recover user data from real gradients collected in a federated learning framework. By minimizing the distance between dummy gradients and the real ones, we can make dummy inputs move closer to the real user data.
Different from the evaluation in the previous gradient attacks~\cite{lu2022april,hatamizadeh2022gradvit,yin2021see}, \emph{we suppose a model should have better capacity of privacy preservation if the data recovered from its gradient updates is less similar to the ground truth input data}. Essentially, we can use the metrics that can reveal the differences or similarities between the input and recovered results. Fortunately, there has already been plenty of previous research in this field. We make use of the off-the-shelf popular metrics for image and text scenarios. Therefore, we use the metrics stated above for evaluation.

\begin{table}[!t]
  \centering
  \caption{Comparisons of different backbones on ImageNet-1K classification. The image sizes here are all set to 224x224.}
  \vspace{-1em}
  \setlength{\tabcolsep}{0.2em} 
  \begin{tabular}{XlccX}
    \toprule
     \textbf{Method} & \textbf{Param.(M)} & \textbf{Top-1 Acc.} $\uparrow$ \\ \midrule 
     ResNet-50~\cite{he2016deep} & 25 & 79.3 \\
     ResNet-50 + MJP & 25 & \textbf{79.4}\\
     DeiT-S~\cite{touvron2021training} & 22 & 79.8 \\
     DeiT-S + MJP & 22 & \textbf{80.5} \\
     Swin-T~\cite{liu2021swin} & 29 & 81.3 \\
     Swin-T + MJP & 29 & \textbf{81.3} \\
     DeiT-B~\cite{touvron2021training} & 86 & 81.8 \\
     DeiT-B + MJP & 86 & \textcolor{black}{\textbf{82.0}} \\
    \bottomrule
  \end{tabular}
  \label{tab:imagenet-pretraining}
\end{table}

\subsection{Image and Text Tasks}
\label{sec:image and text}

\textcolor{black}{This section evaluates MJP purely as a data augmentation technique. We validate its performance across several downstream tasks, including image-related tasks~\ref{subsubsec:image-classification} (\textit{e.g.,} image classification and image segmentation) and text-related~\ref{subsubsec:sentiment-analysis} (\textit{e.g.,} text sentiment analysis and multiple-choice question answering). By setting up experiments outside the context of federated learning, we can isolate the effect of MJP and directly evaluate its standalone contribution.}
\subsubsection{Image Tasks}
\label{subsubsec:image-classification}
\noindent\textbf{Regular ImageNet-1K training.}
For an input image $\mathbf{x}$, we generate its counterpart $\widetilde{\mathbf{x}}$ by applying a Token-wise Jigsaw puzzle to shuffle a portion of selected patches. Besides, a DAL loss has been introduced into the model to strengthen the position information of unshuffled tokens. We mainly compare our model to three typical baselines, including two state-of-the-art Vision Transformers (\emph{i.e.}, DeiT~\cite{touvron2021training} and Swin~\cite{liu2021swin}) and one widely-used CNN-based model, ResNet-50~\cite{he2016deep}. All these baseline models have comparable sizes
). 

\begin{table}[!t]
  \centering
  \caption{Ablation study on the proposed MJP method trained with different masking ratios $\gamma$ on DeiT-S.}
  \vspace{-1em}
  \resizebox{1.\columnwidth}{!}{
  \begin{tabular}{ccccccc}
    \toprule
     \multirow{2}{*}{\textbf{Metric}} & \multicolumn{6}{c}{\textbf{Masking Ratio}}  \\  \cmidrule(lr){2-7} 
     & 0 & 0.03 & 0.09 & 0.15 & 0.21 & 0.27 \\ \midrule
     Top-1 Acc. & 80.0 & \textbf{80.5} & 80.3 & 80.4 & 80.2 & 80.3 \\
    \bottomrule
  \end{tabular}
  \label{tab:ablation-MJP-ratio-cv}}
  \vspace{-1em}
\end{table}

We use the ratio of $\gamma=0.03$ to create $\widetilde{\mathbf{x}}$ in Table~\ref{tab:imagenet-pretraining}. According to these results, the proposed MJP does not have a \textit{negative} effect on the Top-1 accuracy. To our surprise, \textcolor{black}{MJP even achieves a marginal improvement for DeiT-S and DeiT-B}. Besides, MJP works well in the variants of ViTs (\emph{e.g.}, Swin~\cite{liu2021swin}), which shows its potential for generalization ability. Note that we incorporate MJP to ResNet50 by only shuffling the image patches with Alg.~\ref{alg:1} (without PEs). For Swin-T, we apply MJP to the absolute PEs of Swin-T. The detailed experiments and analysis on vision tasks are shown in paper~\cite{ren2023masked}.

\noindent\textbf{Results with different MJP ratios.}
We test different masking ratios used in the Token-wise Jigsaw puzzle strategy during the training. As shown in Table~\ref{tab:ablation-MJP-ratio-cv}, a low ratio $\gamma$ (\textit{e.g.},$0.03$) is sufficient to boost Top-1 Acc. 
As shown in Fig.~\ref{fig:masking-ratio} (a) and (b), 
the performances of original models that are trained without MJP drop dramatically when the masking ratio of the testing samples increases. 
Similarly, even if these models are equipped with MJP with the masking ratio of $\gamma=0.03$, the performance still drops, but at a much slower rate. 
This demonstrates that our proposed MJP method shows more consistent performances. In addition, we also test models trained with different masking ratios. The results show that the model trained with a larger masking ratio demonstrates better robustness to different inference ratios, and the model trained with a larger $\gamma$ is inclined to be more consistent and have less performance drop when the inference ratio increases, as shown in Fig.~\ref{fig:masking-ratio} (c).

\begin{table}[!h]
  \centering
  \caption{Segmentation results on ADE20K dataset (Pre-trained on ImageNet-1K).}
  \vspace{-1em}
  \begin{tabular}{lrrr}
    \toprule
    \textbf{Method} & \textbf{Top-1 Acc.} & \textbf{mIoU} & \textbf{mAcc}\\ \midrule
    Swin-Tiny~\cite{liu2021swin_rebuttal} & 81.3 & 43.87 & 55.22 \\
    Swin-Tiny + MJP & 81.3 & 44.03 (+0.16) & 55.50 (+0.28) \\
    \bottomrule
  \end{tabular}
  \label{tab:segmentation}
  \vspace{-1em}
\end{table}

\begin{figure*}[t]
    \centering
    \begin{subfigure}[b]{0.30\textwidth}
        \centering
        \includegraphics[width=1\linewidth]{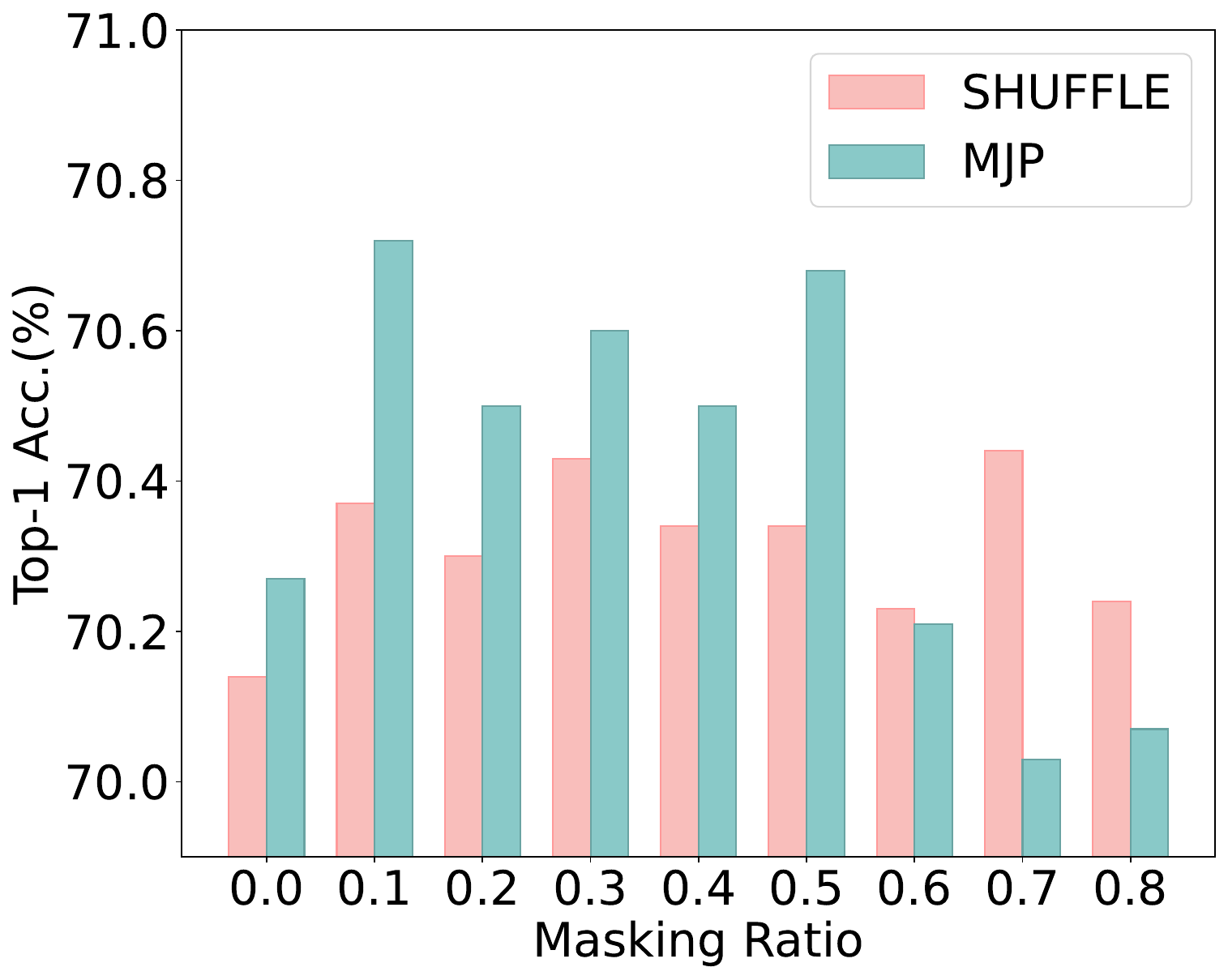}
        \vspace{-6mm}
        \caption{Training with different ratios}
    \end{subfigure}%
    \quad
    \begin{subfigure}[b]{0.30\textwidth} 
        \centering
        \includegraphics[width=1\linewidth]{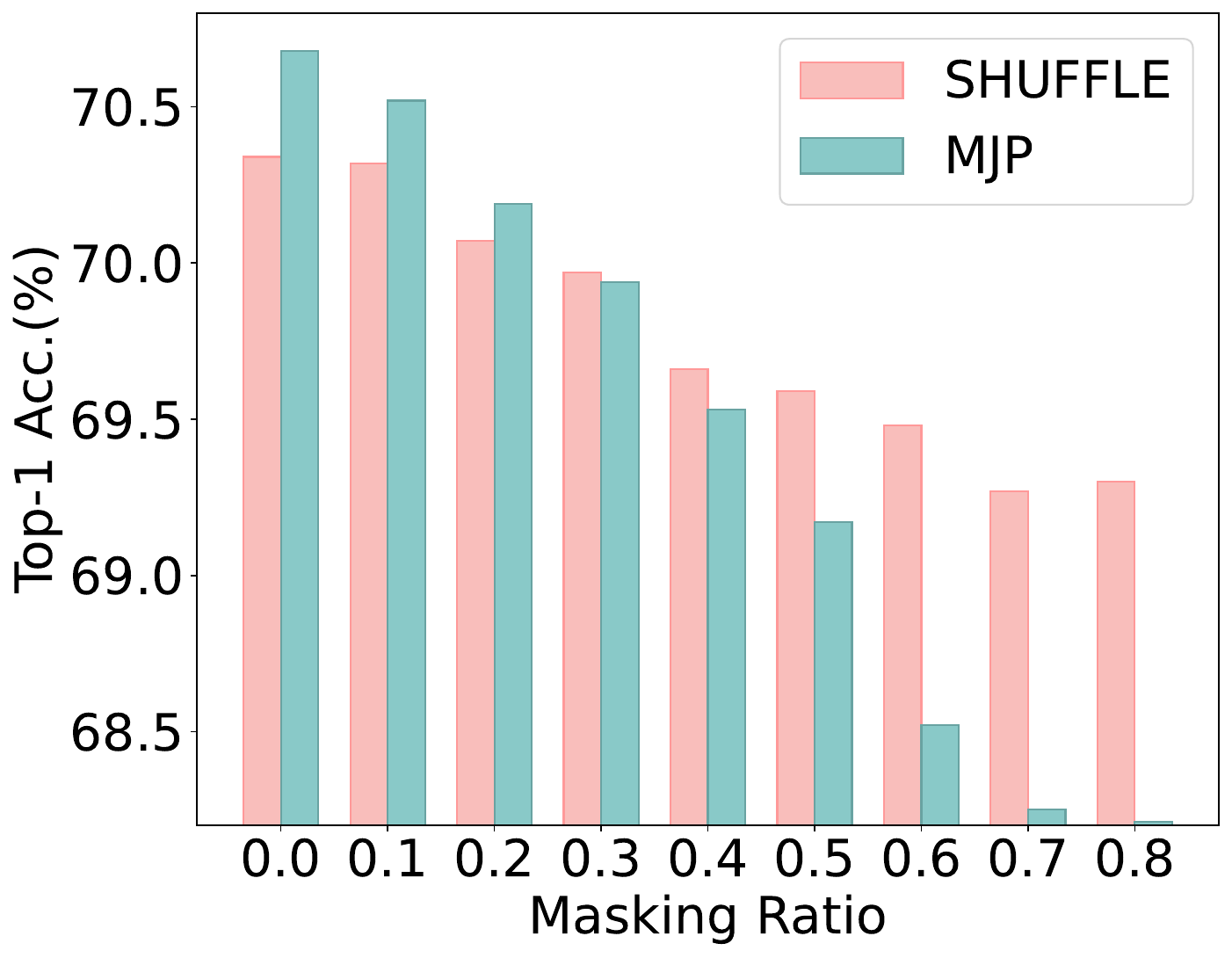}
        \vspace{-6mm}
        \caption{Inference with different ratios}
    \end{subfigure}%
    \quad
    \begin{subfigure}[b]{0.30\textwidth} 
        \centering
        \includegraphics[width=1\linewidth]{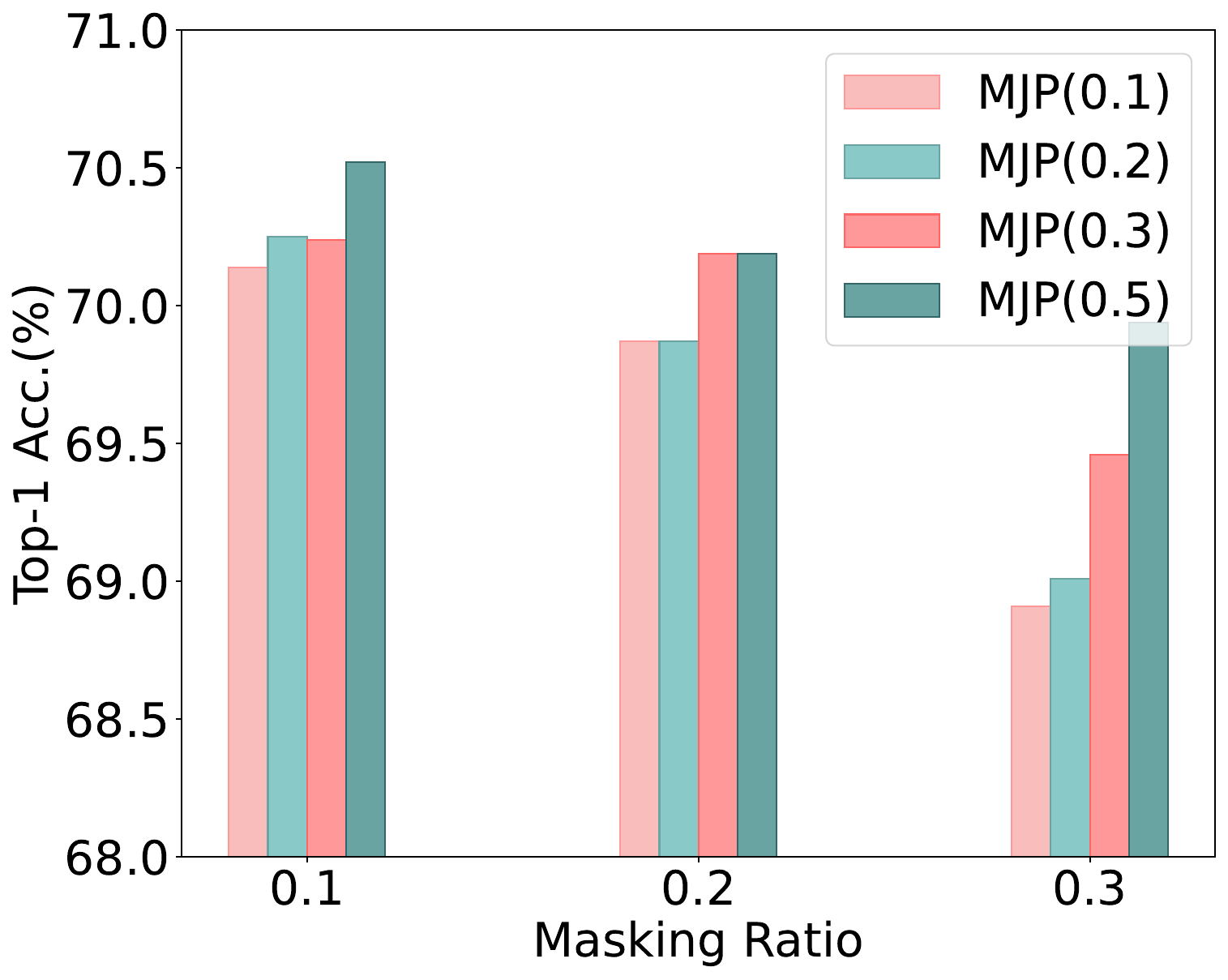}
        \vspace{-6mm}
        \caption{Training with fixed ratio}
    \end{subfigure}
    \vspace{-2mm}
    \caption{Ablation studies over the masking ratio $\gamma$ on the Yelp: (a) comparisons between the \textit{shuffle} and \textit{MJP} method on BERT$_\text{BASE}$ (under different training ratios $\gamma$); (b) comparisons between the \textit{shuffle} and our method on BERT$_\text{BASE}$ (trained with fixed ratio $\gamma=0.5$) when evaluated with different inference masking ratios. (c) comparisons of our method on BERT$_\text{BASE}$ trained with different ratios.}
    \label{fig:masking-ratio-nlp} 
    \vspace{-5mm}
\end{figure*}

\noindent\textbf{Generalization of MJP in Semantic Segmentation Tasks.} To evaluate the generalization capabilities of the proposed MJP method, we conduct experiments on the task of semantic segmentation using the UperNet architecture~\cite{xiao2018unified}, which has been pre-trained with a Swin-Tiny backbone~\cite{liu2021swin_rebuttal} on the ImageNet-1K dataset. The MJP method is integrated into the pipeline by enhancing the feature representations extracted from the Swin-Tiny backbone.These enhanced features are then passed through the standard segmentation head of the UperNet architecture. For evaluation, we use common semantic segmentation metrics, including Top-1 Accuracy, Mean Intersection over Union (mIoU), and Mean Accuracy (mAcc). A higher value for these metrics indicates better model performance, with mIoU measuring the overall overlap between predicted and ground truth segments, and mAcc reflecting the model's average accuracy across different classes.

The results from fine-tuning the Swin-Tiny model with and without MJP on the ADE20K~\cite{zhou2019semantic,zhou2017scene} dataset are shown in Table~\ref{tab:segmentation}. \textcolor{black}{Specifically, when the model introducing MJP is compared to the original Swin-Tiny, we observe a slight increase of 0.16 in mIoU and 0.28 in mAcc.} This indicates that our MJP method does not negatively affect other position-sensitive tasks. These findings highlight that MJP not only preserves but potentially enhances performance in semantic segmentation, demonstrating its robustness across different image understanding tasks.

\begin{table}[!h]
\centering
    \caption{Comparisons of different backbones on Yelp and Amazon datasets for sentiment analysis. Note that \emph{raw} means shuffle ratio $\gamma$ is 0.0, and other $\gamma$ is 0.5. The best value is in \textbf{bold}.}
    \vspace{-1em}
    \setlength{\tabcolsep}{0.1em} 
    \label{tab:nlp-pretraining}
    \resizebox{1.0\columnwidth}{!}{
    \begin{tabular}{XlccX} 
    \toprule
    \textbf{Method} & \textbf{Top-1 Acc.(Yelp)} $\uparrow$ & \textbf{Top-1 Acc.(Amazon)} $\uparrow$ \\
    \midrule
    Transformer4 (raw) & 57.89 & 60.96 \\
    Transformer4 + shuff & 58.00 & 60.99 \\
    Transformer4 + MJP & \textbf{58.05} & \textbf{61.04} \\
    BERT$_\text{BASE}$ (raw) & 70.14 & 71.71 \\
    BERT$_\text{BASE}$ + shuff & 70.34 & 72.03 \\
    BERT$_\text{BASE}$ + MJP & \textbf{70.68} & \textbf{72.17} \\
    \bottomrule
    \end{tabular}
    }
    \vspace{-1em}
\end{table}

\begin{table*}[ht]
  \centering
  \caption{Ablation study on the proposed MJP method trained with different masking ratios ($\gamma$) on BERT$_\text{BASE}$.}
  \vspace{-1em}
  \small
  \setlength{\tabcolsep}{8.5pt}
  \begin{tabular}{lcccccccccc}
    \toprule
    \multirow{2}{*}{\textbf{Metric}} & \multirow{2}{*}{\textbf{Dataset}} & \multicolumn{8}{c}{\textbf{Masking Ratio}} \\ \cmidrule(lr){3-11} 
    & & 0 & 0.1 & 0.2 & 0.3 & 0.4 & 0.5 & 0.6 & 0.7 & 0.8 \\ \midrule
    Top-1 Acc. &Yelp & 70.27 & 70.72 & 70.50 & 70.60 & 70.50 & \textbf{70.68} & 70.21 & 70.03 & 70.07 \\
    &Amazon & 71.82 & 71.83 & 72.00 & 71.99 & \textbf{72.20} & \textbf{72.17} & \textbf{72.01} & 71.85 & 71.84 \\
    \bottomrule
  \end{tabular}
  \label{tab:diffent-MJP-ratio}
  \vspace{-1em}
\end{table*}

\subsubsection{Text Tasks}
\label{subsubsec:sentiment-analysis}

\begin{table}[!h]
  \centering
  \caption{Comparisons study on the different window sizes of the proposed MJP with BERT$_\text{BASE}$ ($\gamma=0.5$). }
  \vspace{-1em}
  \resizebox{1.0\columnwidth}{!}{
  \setlength\tabcolsep{3pt} 
  \begin{tabular}{lllccc}
    \toprule
     \multirow{2}{*}{\textbf{Metric}} & \multirow{2}{*}{\textbf{Method}} & \multicolumn{4}{c}{\textbf{Window Size}}\\  \cmidrule{3-6}
     & & 16 & 32 & 64 & 128   \\  \midrule
      Top-1 Acc.&BERT$_{\text{BASE}}$ + shuff & 70.40 & 70.34 & 70.15 & 70.42  \\ 
      &BERT$_{\text{BASE}}$ + MJP & 70.38 & \textbf{70.68} & 70.56 & 70.25  \\
    \bottomrule
  \end{tabular}
  }
  \label{tab:ablation-window-size}
  \vspace{-1em}
\end{table}

\noindent\textbf{Regular Yelp and Amazon training.}
\textcolor{black}{For text models, we mainly compare with two different training modes, including native shuffle (\textit{shuff}) and \textit{MJP}, which are trained on Transformer4~\cite{vaswani2017attention, deng2021tag} and BERT$_\text{BASE}$~\cite{devlin2018bert}.} Note that we adapt the MJP method as shown in Algorithm~\ref{alg:1} by introducing an $n$-gram window to Transformer4 and BERT$_\text{BASE}$. 
We use $\gamma=0.5$ to create $\widetilde{\mathbf{x}}$ in Table~\ref{tab:nlp-pretraining} (\textit{raw} represents the baseline model using naive shuffle with $\gamma=0.0$). As shown in Table~\ref{tab:nlp-pretraining}, the results consistently demonstrate the performance improvements brought by the MJP method on both the Yelp and Amazon datasets. Compared to the baseline methods (\textit{raw} and naive shuffle), MJP achieves higher Top-1 Accuracy across different datasets and backbones, indicating its strong potential for generalization.

\noindent\textbf{Results with different MJP ratios.}
We also evaluate various masking ratios in the token-wise masking strategy, and all models are trained in the modes of \textit{shuffle} and \textit{MJP} on Yelp. As illustrated in Fig.~\ref{fig:masking-ratio-nlp} (a), according to the performance of model trained with $\gamma > 0$, it shows our method is not sensitive to different $\gamma$.
Notably, a small ratio (\emph{e.g.}, $\gamma = 0.1$) is sufficient to boost accuracy. Additionally, for lower masking ratios (\emph{e.g.}, from 0.0 to 0.5), the model trained with the \textit{MJP} method consistently outperforms the one trained with the \textit{shuffle} method. However, higher masking ratios have a detrimental effect on Top-1 Acc, likely due to the disruption of original sentence semantics caused by excessive shuffling.
Besides, we evaluate the model trained with a fixed training ratio $\gamma=0.5$ by applying different inference masking ratios. As depicted in Fig.~\ref{fig:masking-ratio-nlp} (b), lower ratios (ranging from 0.0 to 0.3) have superior generalization capabilities and better consistency. Conversely, increasing the value of $\gamma$ does not lead to enhanced consistency. Meanwhile, we evaluate models trained with the \textit{MJP} method using different inference masking ratios, as shown in Fig.~\ref{fig:masking-ratio-nlp} (c). In addition, we also observe that when shuffling input tokens at the ratio of 0.5, our proposed models have the optimal generalization capabilities compared with the rest.

\noindent\textbf{MJP ratios across different datasets.}
We also train models using various masking ratios on the Amazon dataset, with results presented in Table~\ref{tab:diffent-MJP-ratio}. We can find that these models exhibit optimal performance at the masking ratio of 0.5, which is similar to the experiments on Yelp. This consistency across different datasets indicates the robustness of our chosen hyper-parameters, (\emph{i.e.}, $\gamma=0.5$). Therefore, we adopt $\gamma=0.5$ for subsequent experiments.


\noindent\textbf{Comparison of different window sizes.}
To investigate the impact of window size $w$ (see Sec.~\ref{subsec:text-models}) on model performance, we experiment with different window sizes, specifically $[16, 32, 64, 128]$, on the Yelp dataset, as detailed in Table~\ref{tab:ablation-window-size}. Based on the average accuracy and the range of accuracy fluctuations, the \textit{shuffle} method achieves 70.33\% accuracy, whereas the \textit{MJP} method slightly outperforms it with an accuracy of 70.47\%. This comparison indicates the superiority of the MJP method over the conventional \textit{shuffle} approach. 
Overall, the performance of the model is less sensitive to the window size, and a window size of 32 demonstrates the highest Top-1 Acc among the tested sizes.

\begin{figure}[!htb]
    \centering
    \includegraphics[width=0.8\linewidth]{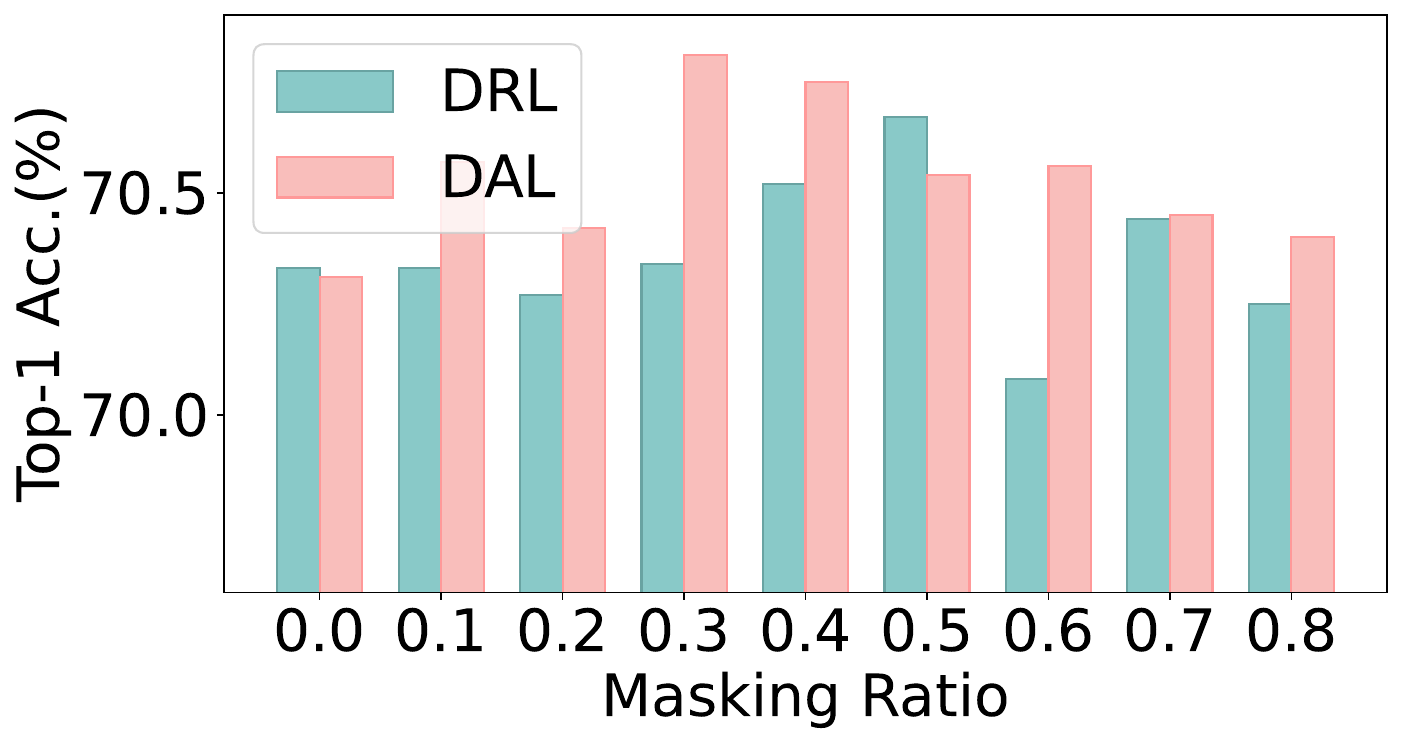} 
    \vspace{-2.5mm}
    \caption{Fine-tuning BERT$_\text{BASE}$ introducing DAL and DRL loss function with different masking ratios.} 
    \label{fig:drlo_different_ratio} 
    \vspace{-2mm}
\end{figure}

\noindent\textbf{Comparison of Auxiliary Loss in NLP.}
Similar to vision tasks, we also explore the effect of the DAL and DRL loss constraints as self-supervised methods on the position embedding of unshuffled tokens in NLP tasks. We train BERT$_\text{BASE}$ with different masking ratios on Yelp, as depicted in Figure~\ref{fig:drlo_different_ratio}. The overall performance comparison between the DAL and DRL loss functions indicates improved results when masking ratios ranges from 0.3 to 0.5, which aligns well with the performance of the MJP method without auxiliary loss (see Table~\ref{tab:diffent-MJP-ratio}). Overall, DAL achieves slightly better accuracy than DRL, but both positional loss functions have little influence on the performance of MJP method when applied to sentiment analysis. 

\noindent\textbf{Generalization of MJP in Multiple-choice QA.}
In addition to sentiment analysis, we extend our evaluation to multiple-choice question answering (QA) using the SWAG dataset~\cite{zellers2018swagaf}, a well-established benchmark for sentence completion and QA tasks. \textcolor{black}{Our experimental results reveal that the MJP method outperforms other approaches (\textit{shuff} and \textit{DRL}), achieving the highest average accuracy (\textit{i.e.,} 81.31\%) across different masking ratios, as shown in Table~\ref{tab:qa-results}. }This reinforces the versatility and robustness of MJP, demonstrating its superior performance not only in sentiment analysis but also in complex, multiple-choice QA tasks. This further expands the applicability of our method across a wider range of NLP tasks, showcasing its generalization capabilities and effectiveness in real-world applications.
\begin{table}[!h]
\caption{Ablation study (\textbf{Top-1 Accuracy}, \%) of the proposed MJP method trained with different masking ratios ($\gamma$) using BERT$_\text{BASE}$ for the multiple-choice QA task on the SWAG dataset.}
\vspace{-1em}
    \centering
    \setlength{\tabcolsep}{3.1pt}
    \begin{tabular}{lcccccccccc}
        \toprule
        Method & 0.0 & 0.1 & 0.2 & 0.3 & 0.4 & 0.5 & 0.6 & 0.7 & 0.8 \\
        \midrule
        shuff & 81.30 & 81.30 & 81.30 & 81.13 & \textbf{81.35} & 81.29 & 81.05 & 81.15 & 81.12 \\
        MJP & 81.26 & 81.47 & 81.30 & \textbf{81.36} &\textbf{81.34} &\textbf{81.35} & 81.12 & 81.32 & 81.23 \\
        DRL & 81.21 & 81.23 & 81.29 & 81.28 &\textbf{81.41} & 81.26 &\textbf{81.38} & 81.28 & 81.24 \\
        \bottomrule
    \end{tabular}
    \label{tab:qa-results}
    \vspace{-1em}
\end{table}

\subsubsection{Effects of MJP on PEs}
Based on the experiments described above, we have demonstrated the effectiveness of the MJP method in enhancing model performance across both two-dimensional images and one-dimensional text modalities. \textcolor{black}{Our method functions as a data augmentation technique, where the order of tokens in the input data is randomly shuffled during training. Given the intrinsic differences in modality/data structure,  task sensitivity and positional encoding mechanisms between text and image data, we outline several key analyses to explain their distinct behaviors under the MJP framework}. 

\noindent\textbf{(1) Masking ratios $\gamma$.} The MJP framework introduces a single new hyperparameter $\gamma$, which balances the trade-off between model performance and generalization ability (\textit{i,e, feature retention}). As illustrated in Table~\ref{tab:ablation-MJP-ratio-cv}, for the image modality, a smaller value (\textit{e.g.}, 0.03) is preferred to improve the performance of the models.
In contrast, for the text modality, a relatively larger $\gamma$ value (\textit{e.g.}, 0.5, compared to 0.03) leads to better performance, which indicates the remarkable inconsistency across these tasks. \textcolor{black}{This discrepancy can be attributed to the difference in modality-specific characteristics and task sensitivity between image and text data.Vision tasks exhibit a increased sensitivity to spatial coherence preservation, thereby requiring minimal shuffle ratios to maintain performance integrity. Conversely, NLP tasks demonstrate greater robustness to contextual disruption due to their reliance on key semantic representations and global context within sentences, enabling the utilization of substantially higher shuffle ratios while sustaining comparable performance metrics.} Therefore, it is crucial to set the value of $\gamma$ below a certain threshold in order to make sure that only an appropriate portion of tokens are perturbed.
This proportion should not be too large. Otherwise, it may cause the loss of excessively sensitive information and lead to inferior performance.

\begin{figure}[!t]
\begin{subfigure}[b]{0.24\textwidth}
        \includegraphics[scale=0.25]{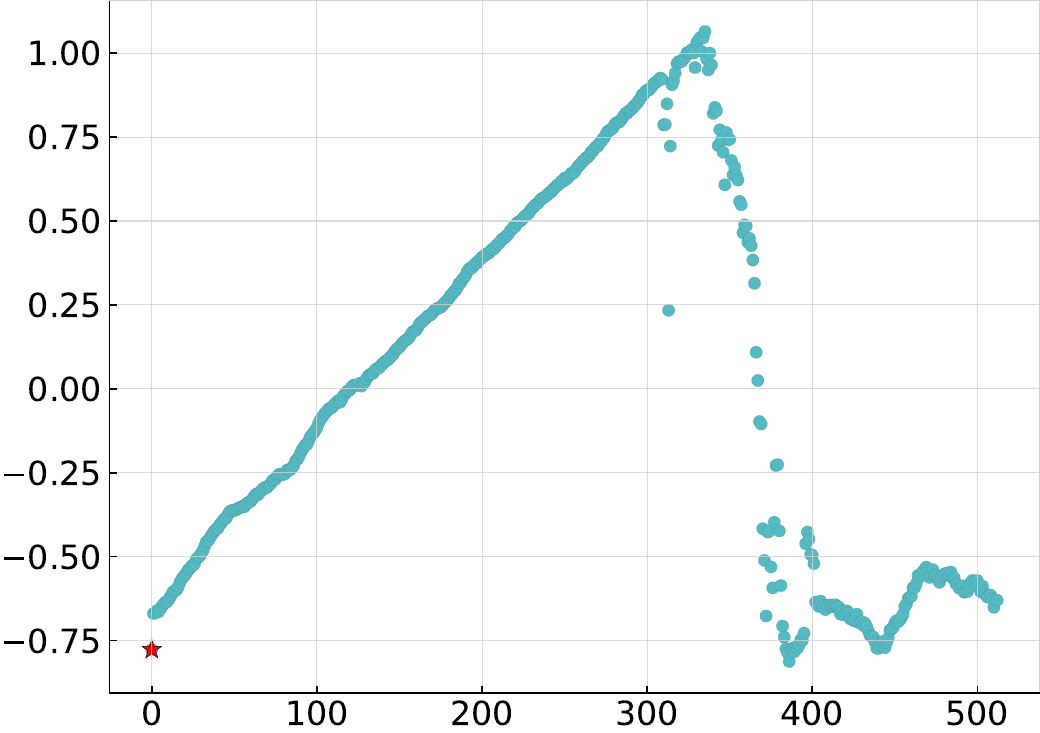}
        \vspace{-6mm}
        \caption{}
    \end{subfigure}  
    \begin{subfigure}[b]{0.24\textwidth}
        \includegraphics[scale=0.25]{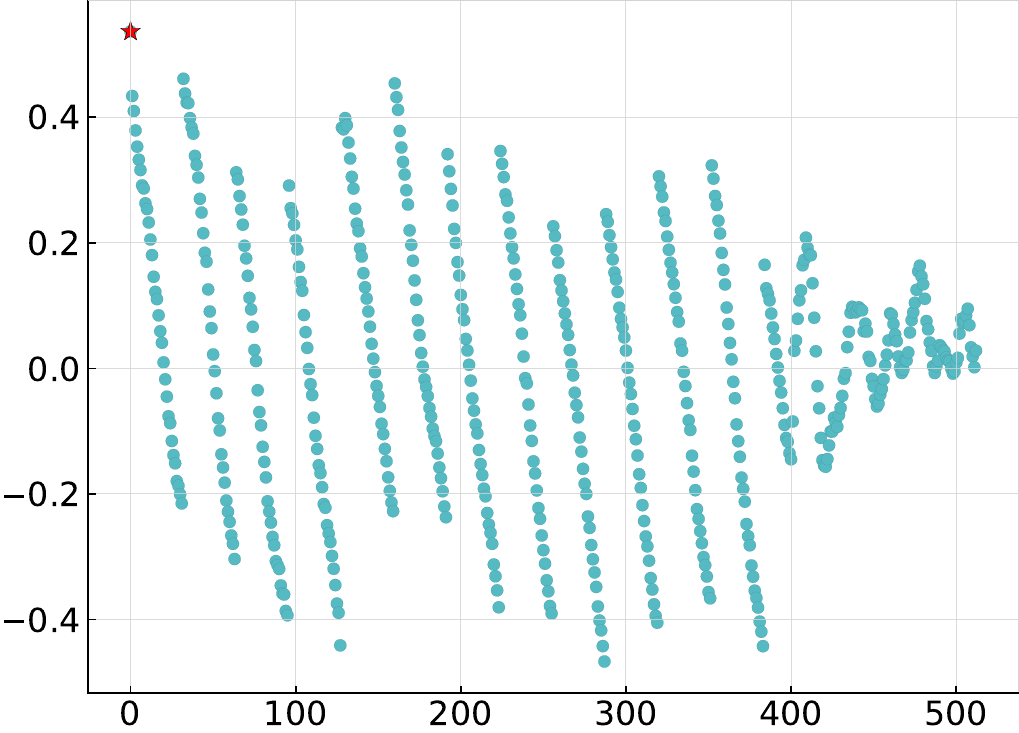}
        \vspace{-6mm}
        \caption{}
    \end{subfigure}  
    \label{fig:mjp_bert_posemb}
    \vspace{-2mm}
    \caption{Low-dimensional projection of position embeddings Fine-tuning BERT$_\text{BASE}$ with MJP. (a) MJP introduces DAL loss function with $\gamma=0.5$. (b) MJP introducing DRL loss function with $\gamma=0.5$.} 
    \label{fig:pos_emb_visualize} 
    \vspace{-2mm}
\end{figure}

\noindent\textbf{(2) PEs processing.} Additionally, for vision tasks, given that the positions of the shuffled patches are two-dimensional, we enhance the MJP framework by introducing a low-dimensional prior for unshuffled patches. This prior is enforced through the proposed DAL regression method, as detailed in Section ~\ref{subsec:dal}, which strengthens the spatial relationships (see line 4 of Alg~\ref{alg:2}). For text tasks, a similar low-dimensional prior has also been implemented (corresponding DAL and DRL loss described in Sec.~\ref{subsec:text-models}), but experimental results~\ref{subsubsec:sentiment-analysis} indicate that this modification has little effect on text tasks.
Additionally, we visualize the PEs of the fine-tuned BERT$_\text{BASE}$ in a 1-dimensional manner, as depicted in Fig.~\ref{fig:pos_emb_visualize}. The red star highlights the first position token, and the maximal position index is limited to 512.
Due to the sparsity of long sentences, marks at distant positions (approximately 320-400) appear more scattered and randomly distributed.
This is probably because the majority of our selected sentences have the length less than 320. 
As shown in Fig.~\ref{fig:origin_posemb} (c), the visualization of the original PEs for BERT$_\text{BASE}$ has the shape of sinusoidal curve.
When adding the DAL constraint, the sinusoidal curve is transformed into a linear shape as shown in Fig.~\ref{fig:pos_emb_visualize} (a), and it is transformed into a neatly arranged bar pattern in Fig.~\ref{fig:pos_emb_visualize} (b) when using DRL loss. These experimental results suggest that the positions of the original tokens can be effectively represented in a low dimension.

\begin{table}[!t]
  \centering
  \caption{Comparisons on gradient leakage by analytic attack~\cite{lu2022april} with ImageNet-1K validation set, where we test (1) ViT-S, DeiT-S and our model in the setting (a); (2) ViT-S, DeiT-S and our model in the setting (b) (\emph{i.e.}, MJP with $\gamma=0.27$); (3) ablation on without (w/o) using $\mathbf{E}_{\text{unk}}$ in setting (a); and (4) Our model in setting (c).}
  \vspace{-1em}
  \resizebox{1.0\columnwidth}{!}{
  \setlength{\tabcolsep}{0.15em} 
  \begin{tabular}{XllcccccccX}
    \toprule
     & \textbf{Model} &  \textbf{Set.} & \textbf{Acc.} $\uparrow$  & \textbf{MSE} $\uparrow$ & \textbf{FFT$_\text{2D}$} $\uparrow$ & \textbf{PSNR}~$\downarrow$ & \textbf{SSIM} $\downarrow$ & \textbf{LPIPS} $\uparrow$ 
     \\ \midrule 
     \multirow{2}{*}{(1)} & ViT-S~\cite{dosovitskiy2020image} & \multirow{4}{*}{a} & 78.1 &  .0278 & .0039 & 19.27 & .5203 & .3623 \\
     & DeiT-S~\cite{touvron2021training} & & 79.8  & .0350 & .0057 & 18.94 & .5182 & .3767 \\
     & DeiT-S (w/o PEs) & & 77.5 & .0379 & .0082 & 20.22 & .5912 & .2692 \\
     & DeiT-S+MJP & & \textbf{80.5}  & \textbf{.1055}  & \textbf{.0166} & \textbf{11.52} & \textbf{.4053} & \textbf{.6545} 
     \\ \midrule
     \multirow{4}{*}{(2)} & ViT-S~\cite{dosovitskiy2020image} & \multirow{4}{*}{b} & 18.7 &   .0327 & .0016 & 18.44 & .6065 & .2836 \\
     & DeiT-S~\cite{touvron2021training} & & 36.0 & .0391 & .0024 & 17.60 & .5991 & .3355 \\ 
     & DeiT-S (w/o PEs) & & \textbf{77.5} & .0379 & .0025 & 20.25 & .6655 & .2370 \\
     & DeiT-S+MJP & & 62.9 &   \textbf{.1043} & \textbf{.0059} & \textbf{11.66} & \textbf{.4493} & \textbf{.6519} 
     \\ \midrule
     (3) & DeiT-S+MJP (w/o) & a  &  40.6 & .1043 & .0059 & 11.66 & .4493 & .6519 
     \\ \midrule
     (4) & DeiT-S+MJP & c & 62.9   &  .1706 & .0338 & 8.07 & .0875 & .8945 \\ \bottomrule
  \end{tabular}}
  \label{tab:grad-attack-image}
  \vspace{-1em}
\end{table}

\subsection{Preserving User Privacy}
\label{sec:Preserving User Privacy}
\textcolor{black}{To evaluate the privacy-preserving capabilities of MJP, we integrate it into a federated learning setup and evaluate it against gradient inversion attacks. The core principle of these attack methods is that each sample activates only a portion of content-related neurons in deep neural networks, leading to one specific backward gradient for one related sample (\emph{i.e.}, 1-to-1 mapping). Based on this observation, we argue that feeding a Transformer model with permuted input tokens can intuitively disrupt this mapping, thereby misleading the attack, as shown in Fig. ~\ref{fig:privacy_attacker} (Original path \textit{vs.} MJP path). To validate this quantitatively, we simulate a privacy attack scenario where an adversary attempts to reconstruct original input data from shared gradient updates.}
The attack experiment is set in three modes. Given an image/text $\mathbf{x}$ and its transformed counterpart (\emph{e.g.}, token shuffled) $\tilde{\mathbf{x}}$, a transformer-based model $\mathcal{M}$, and automatic evaluation metrics $\phi$,  we naturally conduct three different settings for fair comparisons: (a) $\phi(\nabla{\mathcal{M}(\mathbf{x})}, \mathbf{x})$, (b) $\phi(\nabla{\mathcal{M}(\tilde {\mathbf{x}} )}, \tilde{\mathbf{x}})$, and (c) $\phi(\nabla{\mathcal{M}(\tilde{\mathbf{x}})}, \mathbf{x})$, where $\nabla$ denotes the process of recovering input data through gradient attacks. 
\begin{table}[!t]
  \small
  \centering
  \caption{Comparisons on gradient leakage by optimization-based attack~\cite{lu2022april} with ImageNet-1K validation set, where we test (1) ViT-S and our model in the setting (a); (2) ViT-S and our model in the setting (b); and (4) Our model in setting (c).}
  \vspace{-1em}
  \resizebox{1.0\columnwidth}{!}{
  \setlength{\tabcolsep}{0.15em} 
  \begin{tabular}{XllcccccccX}
    \toprule
     & \textbf{Model} &  \textbf{Set.}  & \textbf{MSE} $\uparrow$ & \textbf{FFT$_\text{2D}$} $\uparrow$ & \textbf{PSNR}~$\downarrow$ & \textbf{SSIM} $\downarrow$ & \textbf{LPIPS} $\uparrow$ 
     \\ \midrule 
     \multirow{2}{*}{(1)} & ViT-S~\cite{dosovitskiy2020image} & \multirow{2}{*}{a} &.0446 &.0176 &14.25  &.3355 &.6159 \\
     & ViT-S+MJP &   &.0442 &.0165 &14.20  &.3364 &.6113 \\
      \midrule
     (2) & ViT-S+MJP & b &.0482 &.0251 &13.72  &.3728 &.6551 \\ 
    \midrule    
     (3) & ViT-S+MJP & c &\textbf{.0482} &\textbf{.0255} &\textbf{13.71}  &\textbf{.3726} &\textbf{.6544} 
     \\ \bottomrule
  \end{tabular}}
  \label{tab:grad-attack-image-opitimazion}
  \vspace{-1em}
\end{table}

\begin{table*}[!ht]
  \centering
  \small
  \caption{Comparisons on gradient leakage by optimization-based attack~\cite{deng2021tag} with Yelp validation set, where we test (1) Transformer4 model in the setting (a); (2) Transformer4 and Transformer4+MJP model in the setting (b); (3) ablation on our model in setting (c); (4) The ablation on our model as increasing the number of iteration.}
  \vspace{-1em}
  \setlength{\tabcolsep}{5.5pt}
  \begin{tabular}{lllcccccccccc}
    \toprule
     & \textbf{Set} &\textbf{Model} & \textbf{Iter} & \textbf{Ratio} & \textbf{Acc} $\downarrow$  
     & \textbf{S-bleu} $\downarrow$ &\textbf{G-bleu} $\downarrow$ & \textbf{Rouge1} $\downarrow$ 
     & \textbf{Rouge2} $\downarrow$ & \textbf{RougeL} $\downarrow$  &\textbf{Token\_ac} \\
    \midrule
    (1) & (a) &Transformer4  &5k &0.0  &41.62 &31.90  &20.86  &39.21  &22.41   &38.89   &41.81  \\
      \midrule
     \multirow{2}{*}{(2)} &(b) &Transformer4   &5k &0.9  &41.67 &32.28  &20.61  &39.68 &22.04  &39.17  &41.80 \\
     & (b) &Transformer4 + MJP  &5k &0.9  &\textbf{0.98} &\textbf{0.63}  &\textbf{0.28}  &\textbf{1.41} &\textbf{0.01}  &\textbf{1.36}  &\textbf{1.21} \\   
     \midrule 
      \multirow{2}{*}{(3)} & (c) &Transformer4 &5k &0.9  &17.33  &29.12  &13.88  &37.50 &10.07  &26.50  &41.80  \\
      & (c) &Transformer4 + MJP  &5k &0.9    &\textbf{0.98} & \textbf{0.63} & \textbf{0.28}  &\textbf{1.40}  &  \textbf{0.01} &\textbf{1.33} & \textbf{1.21} \\
      \midrule
      \multirow{2}{*}{(4)} & (b) &Transformer4 + MJP  &30k &0.9 &2.95 & 2.41 & 0.85  &  3.72  &0.34 &3.65 &3.17 \\
      & (c) & Transformer4 + MJP  &30k  &0.9  &2.95 &2.40  &0.85   &3.72  &0.34 &3.63  &3.17 \\
     \bottomrule
  \end{tabular}
  \label{tab:grad-nl-attack}
  \vspace{-1em}
\end{table*}
\subsubsection{Gradients Inversion on Vision}
\noindent\textbf{Analysis-Based Gradient Inversion.} We utilize the Analytic Attack method proposed in APRIL~\cite{lu2022april}, which is designed to attack the ViTs by reconstructing images from gradient updates. For this experiment, we randomly sample 1K images from the validation set of ImageNet-1K (\emph{i.e.}, one image per category). Table~\ref{tab:grad-attack-image} presents a quantitative comparison between our method and the original ViTs for batch gradient inversion. In setting (a), ViT-S achieves the best PSNR with the value of 19.27 (shown in the first three rows), and also has competitive performance \emph{w.r.t} other evaluation metrics. This demonstrates that APRIL~\cite{lu2022april} effectively reconstructs original images from the gradient updates of the raw ViTs. However, the introduction of MJP framework (\textit{e.g.,} DeiT-S+MJP) results in a PSNR reduction to 11.52, which indicates a degradation in image recovery quality. This decline extends to other metrics: SSIM falls from 0.5203 to 0.4053, and LPIPS improves significantly from 0.3623 to 0.6545, which suggests an increase in image dissimilarity. These results clearly demonstrate effectiveness of our MJP framework in defending against analysis-based attacks using the Analytic Attack in APRIL.

\begin{figure}[!h]
\vspace{-1em}
 \centering
     \includegraphics[width=1\linewidth]{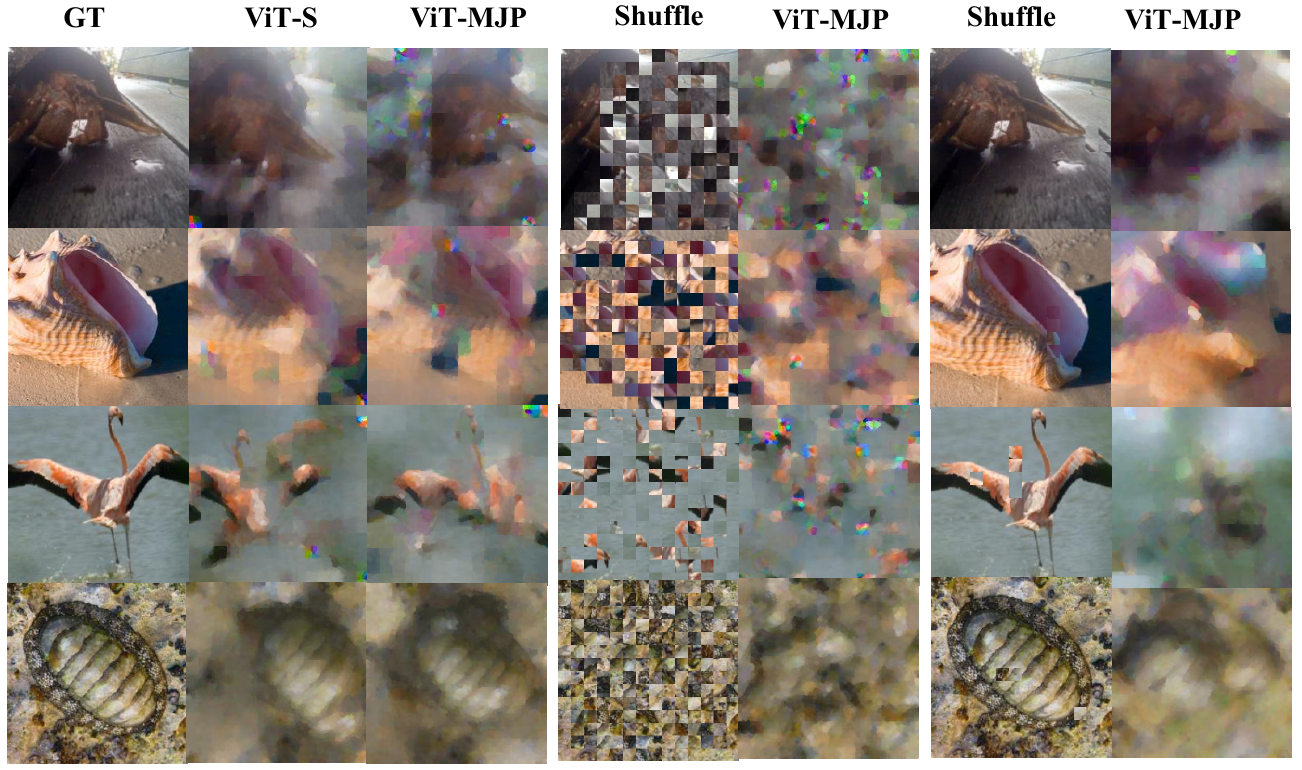}
    \caption{Visual comparisons on image recovery with gradient updates~\cite{hatamizadeh2022gradvit}. Our proposed ViT-S+MJP ($\gamma=0.9$ for column 5, and $\gamma=0.03$ for column 7 ) model outperforms the original ViT-S~\cite{dosovitskiy2020image}.}
    \label{fig:image_attack} 
    \vspace{-5mm}
\end{figure}
\noindent\textbf{Optimized-Based Gradient Inversion.} We further validate the MJP method using an optimization-based gradient inversion approach on ImageNet-1K, following protocols outlined in previous studies~\cite{hatamizadeh2022gradvit, yin2021see}. Despite the absence of publicly available code from the original researchers, we have closely replicated their methods based on the detailed descriptions provided in the literature. As illustrated in Table~\ref{tab:grad-attack-image-opitimazion}, while the MSE for ViT-S is 0.0446 in setting (a), our model achieves slightly higher MSE values of 0.0482 in settings (b) and (c), respectively. Similarily, our model demonstrates improved metrics with PSNR at 13.71, SSIM at 0.3726, and LPIPS at 0.6544, compared to 14.25, 0.3355, and 0.6159 for ViT-S. These results demonstrate that our method consistently outperforms the original ViT-S across all evaluation metrics, which indicates that our approach offers superior privacy protection ability. 
A visual comparison between the results from other methods and ours is presented in Fig.~\ref{fig:image_attack}, where we successfully recover the ground truth with ViT-S (see first and second columns in Fig.~\ref{fig:image_attack}). However, when the MJP method is applied, the recovered images lack any clear details from the original image (see the third ($\gamma=0$), fifth ($\gamma=0.9$) and seventh ($\gamma=0.03$) columns in Fig.~\ref{fig:image_attack}). Although our baseline implementation does not achieve optimal results, these cases demonstrate that MJP offers a certain degree of defense against optimization-based attacks on Transformers.
(Note: \textit{Optimizing the recovery process is not the primary focus of our research due to the absence of the original code from the paper. Instead, we focus on information retained from the original image in the recovered image. Our goal is to verify the effectiveness of MJP in terms of the defense.})


\begin{table*}[!ht]
 \scriptsize
  \centering
  \caption{Recover examples for the text task on Yelp. The ``GT'' denotes the raw input sentence, and the ``Recover'' presents the reconstructed sentence by gradient inversion in different modes. Words that have been reconstructed incorrectly is highlighted in red.}
  \vspace{-1em}
  \begin{tabularx}{\textwidth}{p{0.08\textwidth} p{0.08\textwidth} p{0.02\textwidth} X}
    \toprule
    \textbf{Category} &\textbf{Mode} &\textbf{Iter} &\textbf{Text} \\
    \midrule
    \textbf{GT} &-- &--  &dr. goldberg offers everything i look for in a general practitioner.  he's nice and easy to talk to without being patronizing; he's always on time in seeing his patients; he's affiliated with a top-notch hospital (nyu) which my parents have explained to me is very important in case something happens and you need surgery; and you can get referrals to see specialists without having to see him first.  really, what more do you need?  i'm sitting here trying to think of any complaints i have about him, but i'm really drawing a blank.\\
    \addlinespace
    \textbf{Recover GT($\gamma$=0.0)} &-- &3k &dr. goldberg offers everything i look for in a general practitioner.  he's nice and easy to talk to without being patronizing; he's always on time in seeing his patients; he's affiliated with a top-notch hospital (nyu) which my parents have explained to me is very important in case something happens and you need surgery; and you can get referrals to see specialists without having to see him first.  really, what more do you need?  i'm sitting here trying to think of any complaints i have about him, but i'm really drawing a blank.\\
    \addlinespace

    \midrule
    \textbf{Shuffled GT($\gamma$=0.1)} &-- &-- &dr. goldberg offers everything i look for in a general practitioner.  he's nice and easy to talk to without being patronizing; he's always on time in seeing his patients; he's affiliated with a top-notch hospital (nyu) which my parents have explained to me is very important in case something happens and and need surgery; you you can get referrals to see specialists without having to see him first.  really, what more do you need?  i'm to here trying sitting think of any complaints i have about him, but i'm really drawing a blank.\\
    \addlinespace
    \textbf{Recover($\gamma$=0.1)} &Transformer4 + MJP &100k  & dr. \textcolor{red}{water}berg offers \textcolor{red}{obliter} i look for in a general practitioner.  he's nice and \textcolor{red}{Verse toJ} to without being patronizing; he's always on time\textcolor{red}{expression} seeing his patients; he's affiliated with a top-not\textcolor{red}{Wan hospital (ny Rih) which} my \textcolor{red}{obliter} have explained to me is very important in case something happens and \textcolor{red}{430} need surgery; \textcolor{red}{urd} you can get referrals to see specialists without \textcolor{red}{lions} to see him first.  really, what more do \textcolor{red}{Groups} need? i'\textcolor{red}{mexpression} here trying \textcolor{red}{houses} think of any complaints i have about him, but i'm really drawing a blank.\\
    \midrule
    \addlinespace
    \textbf{Shuffled GT($\gamma$=0.5)} &-- &-- & practitioner adrberg offers. i look everything for gold general in.  he easy to and niceizing to talk without being patron's; he's always on top seeing hospital his patients; affiliated's he with ach-not time in explained tou) which my parents have is ( meny very important in case something specialists happens to need surgery; and you get you referrals and see can without having  to him first more. see, what need do you really?  i'm sitting i trying to think any about complaints, of have him here but drawing'm really i a blank.\\
    \addlinespace
    \textbf{Recover($\gamma$=0.5)} &Transformer4 + MJP &100k & 
    \textcolor{red}{knockingantics water} berg offers \textcolor{red}{obliter} i look \textcolor{red}{baptism Yam solic} general \textcolor{red}{water}.  he \textcolor{red}{customized review} and \textcolor{red}{VerseUSAexpression 430} without being patron \textbf{incorrectly proposed} he\textcolor{red}{friendly} always on \textcolor{red}{Wagnor571rays} patients; \textcolor{red}{soda's proposed} with \textcolor{red}{a incorrectly Caucasnot Luthorexpressionclosureraysu) which faculties parents disapp obliter.) meWan Under important}expression \textcolor{red}{case something FUCKWan deprivation need obliter;ZZins Forumjit referrals drivers blocks Bett without having helpingChicago him first350 flavoredfecture,fourth paradigm do youupdate?  i 430} sitting \textcolor{red}{paradigm} trying to \textcolor{red}{informational Artifact hospitalsWan paradigm AVclosure him diameter butWanexcept really informational Enlightenment} blank. \\
    \bottomrule   
  \end{tabularx}
  \label{tab:data-example}
  \vspace{-1em}
\end{table*}


\begin{figure*}[ht]
    \centering
    \begin{subfigure}[b]{0.27\textwidth}
       \includegraphics[width=\linewidth]{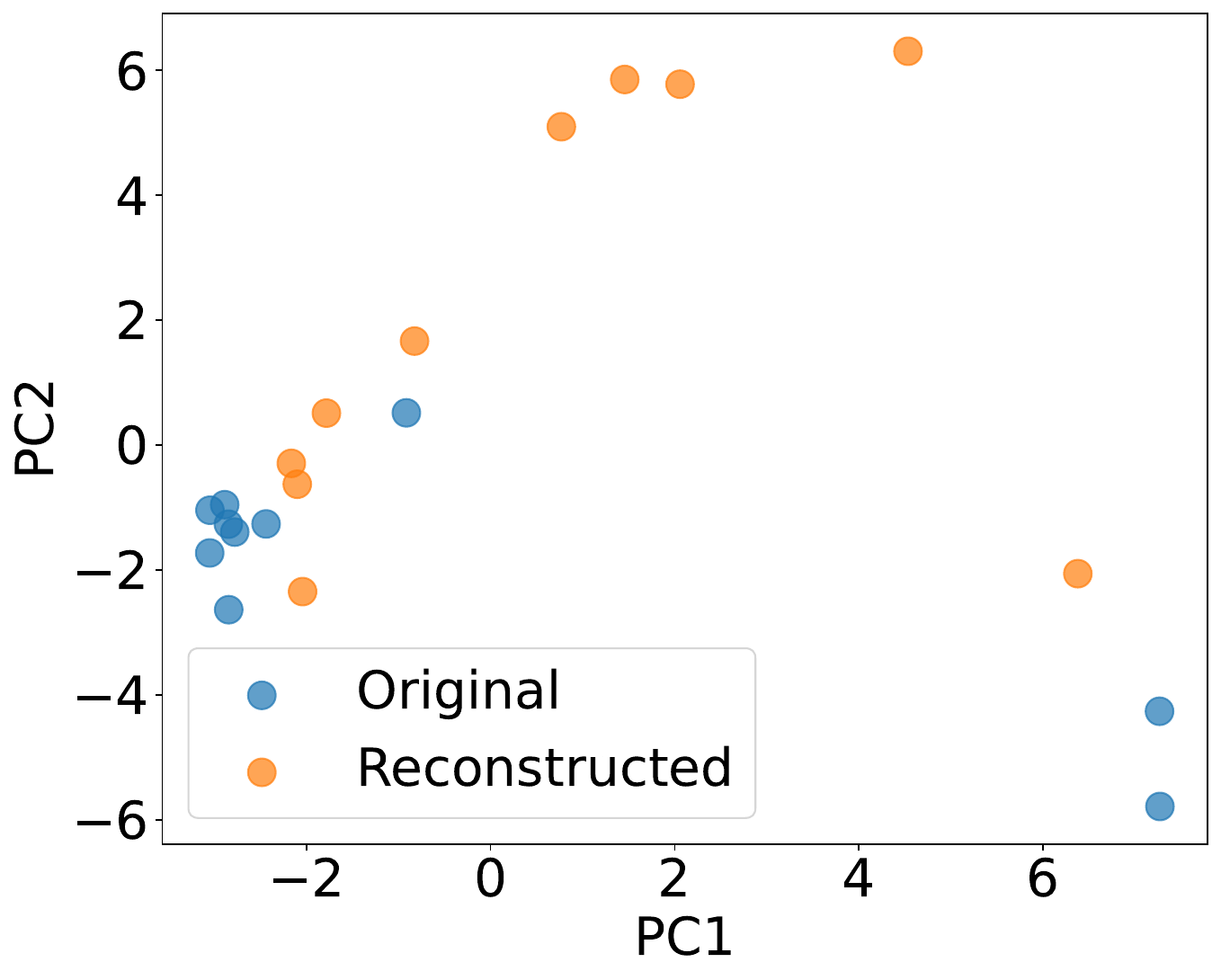}
       \vspace{-6mm}
       \caption*{\hspace{6mm}(a)}
    \end{subfigure}
    \hfill
    \begin{subfigure}[b]{0.27\textwidth} 
        \centering
        \includegraphics[width=\linewidth]{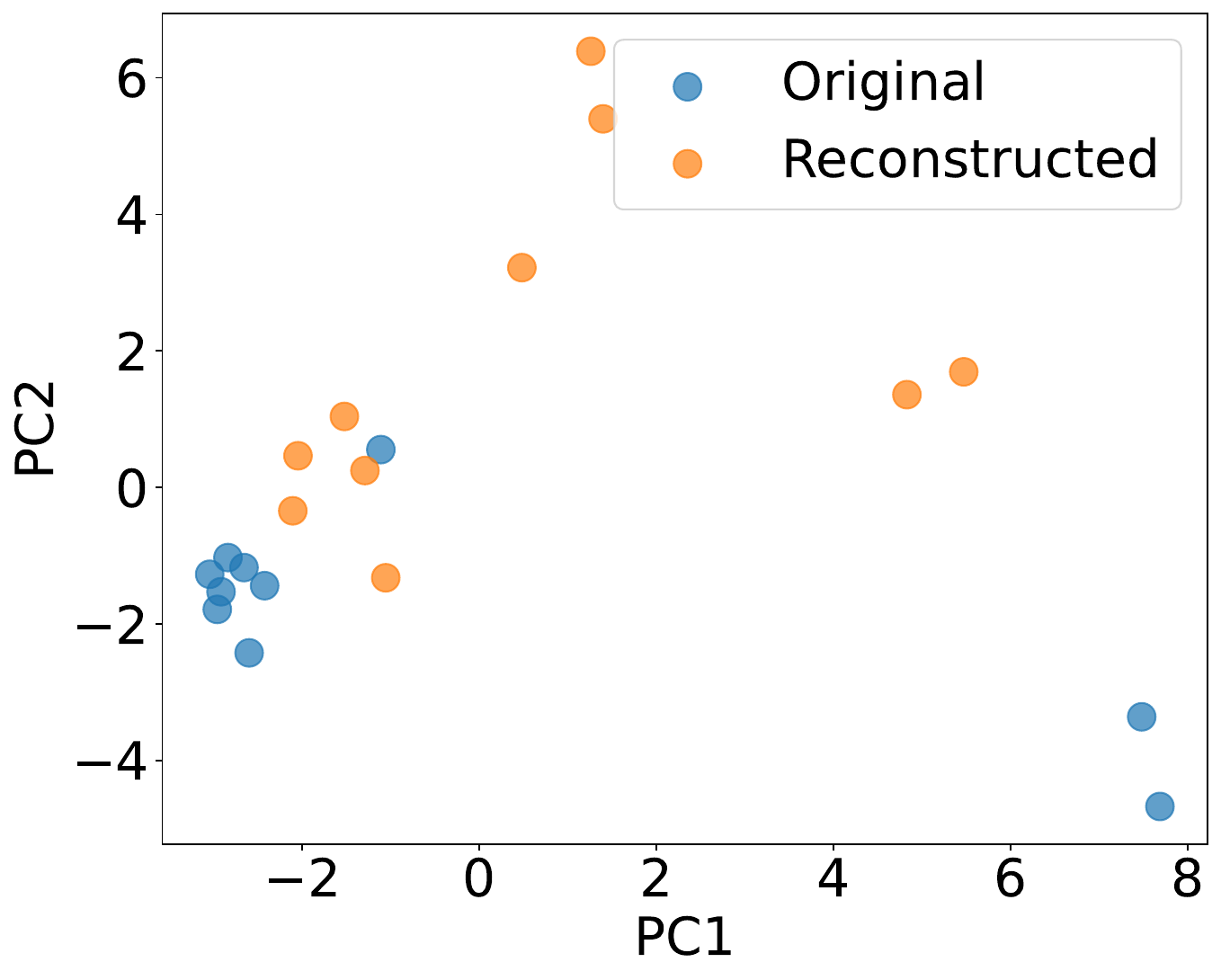}
        \vspace{-6mm}
        \caption*{\hspace{6mm}(b)}
    \end{subfigure}
    \hfill
    \begin{subfigure}[b]{0.27\textwidth} 
        \centering
        \includegraphics[width=\linewidth]{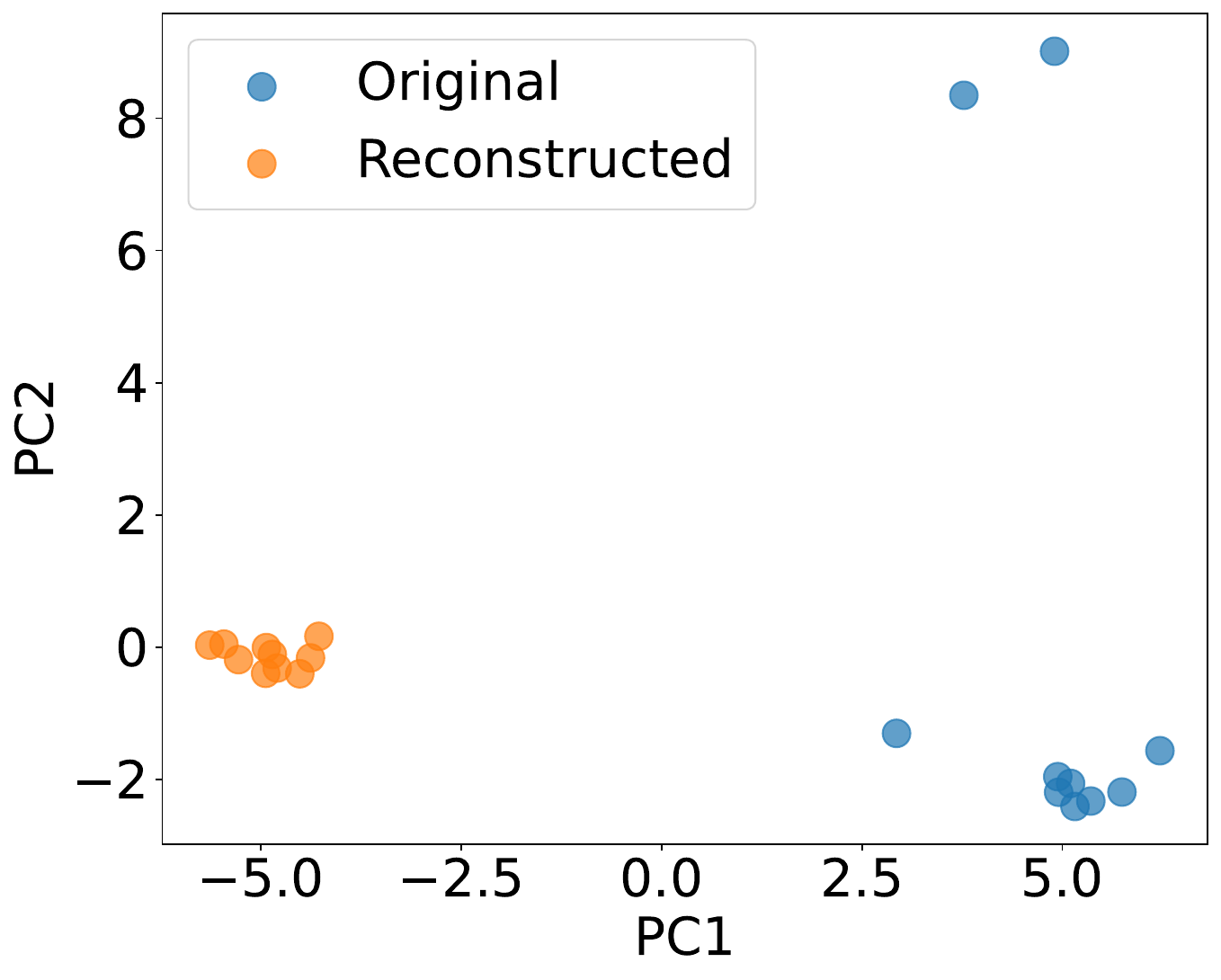}
        \vspace{-6mm}
        \caption*{\hspace{6mm}(c)}
    \end{subfigure}
    \vspace{-2mm}
    \caption{2D PCA plots for the dimension-reduced token embeddings of Transformer4 on Yelp dataset. The cosine similarity of dimension reduced token embeddings between recovered data (\textit{Reconstructed}) and ground truth (\textit{Original}) with different modes and shuffle $\gamma$. (a) Model$_\text{shuffle}$, $\gamma=0.0$. (b) Model$_\text{shuffle}$, $\gamma=0.5$. (c) Model$_\text{MJP}$, $\gamma=0.5$.}
    \label{fig:recover_similarity} 
    \vspace{-5mm}
\end{figure*}

\subsubsection{Gradients Inversion on Text}
For the text task, we validate our algorithm on Transformer4 and BERT$_\text{BASE}$ using the proposed TAG~\cite{deng2021tag}, an optimization-based attack, which enables a viable and complete recovery of the original sentence from the gradient updates of the original transformer. We randomly sample 100 sentences from Yelp validation dataset, and the downstream task of the attack is five-class sentiment analysis. Notably, we are the first to perform recovery based on Transformer models for long texts and to conduct a quantitative analysis. In this experiment, we use $\gamma=0.03$ to create $\widetilde{x}_t$ in the section experiment.

\noindent\textbf{Quantitative Comparisons.} Table~\ref{tab:grad-nl-attack} presents a quantitative comparison between our method and the original Transformer4 for gradient inversion on Yelp. The experimental results show that our proposed method causes TAG to yield unrecognizable sentences and fail to recover the details (\emph{i.e.}, introducing noisy tokens in the outputs). Then, as illustrated in Table~\ref{tab:grad-nl-attack} (row 4), even with an increased number of iterations, MJP is still unable to recover more information from the original sentence. Some cases are provided in Table.~\ref{tab:data-example}. 

\noindent\textbf{Vulnerable Population.} \textit{(1) Data}. We know that shorter sentences (\textit{e.g.,} SST2~\cite{socher-etal-2013-recursive} and CoLA~\cite{warstadt2018neural}) are more easily recovered after only a few hundred of iterations~\cite{deng2021tag}. Here, experimental results show longer sentences (180-400 words) require a greater number of iterations for recovery. \textit{(2) Model Architecture}. Transformer4, with fewer blocks and a smaller word embeddings dimension (96), is more susceptible to attacks than BERT$_\text{BASE}$, which has a word embedding dimension of 768. The increased complexity of BERT$_\text{BASE}$ makes it harder to attack and requires more time for data recovery~\cite{deng2021tag}. Consequently, simpler model structures are more vulnerable and at higher risk of privacy leakage (\emph{i.e.,} more prone to gradient inversion), as demonstrated by the case in Table~\ref{tab:data-example}. \textit{(3) Our Method}. When the MJP method is introduced, data recovery becomes extremely difficult, even increasing the number of iterations to the maximum (\textit{i.e.,} until convergence). This suggests a noticeable improvement in the model's defense capability against gradient inversion. These results indicate that MJP is promising for protecting user privacy in text tasks within federated learning.  

\noindent\textbf{PEs Visualization.} To intuitively assess the similarity of long sentences, we reduce the dimensionality of the ground truth and recovered token embeddings to 2D space using PCA. We then evaluate the similarity of these dimension-reduced token embeddings using the cosine similarity~\cite{2020On}. As shown in Fig~\ref{fig:recover_similarity} (a) and (b), when we use \textit{shuffle} mode, the original and reconstructed sentences are indistinguishable, with all data points intermixed. However, with the introduction of the MJP method, the original and the reconstructed sentences form distinct clusters, which indicates that the reconstructed sentences are significantly different from the original ones. This separation in the low-dimensional space demonstrates that the gradient inversion attack fails to reconstruct token embeddings that are semantically and structurally close to the original input.


%% file: sections/5conclusion.tex
\section{Discussion and Conclusion}
\label{sec:discussion and conclusion}
\subsection{Discussion}
\noindent\textbf{PCA Projected Dimensionality.} 
Explained variance (EV) in PCA represents the proportion of the total variance attributed (explained) by each principal component. Table~\ref{tab:variance_dim} presents EV of our models (DeiT-S+MJP and BERT$_\text{BASE}$+MJP) and baseline models (original DeiT-S~\cite{touvron2021training} and BERT$_\text{BASE}$~\cite{2018BERT}) across different projection dimensions. As shown in Table~\ref{tab:variance_dim}, both DeiT-S and BERT$_\text{BASE}$ models achieve EV percentages exceeding 50\% with fewer than ten dimensions. DeiT-S reaches up to 90.74\% with 7 dimensions, and BERT$_\text{BASE}$ up to 62.35\% with 10 dimensions. \textcolor{black}{This indicates that low dimensionality can capture a significant amount of spatial information, which stems from the structure of learnable PEs as a lookup table, an inherently sparse representation. Furthermore, to achieve the same explained variance ratio, our MJP-based models require more dimensions than the baseline models. This indicates that the position embedding matrix in ours is less sparse but more informative, and it has been validated on both vision and text modalities. To sum up, in vision tasks, the informativeness of the PEs between our vision models and the baseline models exhibits notable differences due to the integration of MJP into the models. 
In contrast, marginal differences are observed in text data, likely resulting from the simpler positional information in 1D text data. Despite these variations, the consistent trend across both modalities validates MJP's effectiveness in learning more informative position representations.}

\begin{table}[!t]
  \centering
  \caption{Explained variance versus PCA projected dimensionality on vision and text models.}
  \vspace{-1em}
  \resizebox{1.0\columnwidth}{!}{
  \begin{tabular}{cccccc}
    \toprule
     \textbf{Projected Dimension} & 3 & 4 & 5 & 6 & 7 \\
    \midrule
    \textbf{DeiT-S EV (\%)} &54.61 & 68.55 & 77.95 & 85.54& 90.74\\
    \textbf{DeiT-S+MJP EV (\%)} &46.74 &58.36 &69.10 &78.13 &84.55 \\
    \midrule
     \textbf{Projected Dimension} & 6 & 7 & 8 & 9 & 10 \\
    \midrule
    \textbf{BERT EV (\%)}  &43.78 &49.42 &53.84 &58.15 &62.35\\
    \textbf{BERT+MJP EV (\%)} &43.32 &48.87 &53.31 &57.55 &61.70 \\
    \bottomrule
  \end{tabular}
  \label{tab:variance_dim}
  }
  \vspace{-1em}
\end{table}

\noindent\textbf{Accuracy \emph{Vs.} Shuffle Ratio.}
Intuitively, increasing the shuffle ratio with the proposed MJP method might seem to undermine the original intrinsic inductive bias. However, experimental results demonstrate that the performance of Transformer models is not only preserved but also actually boosted via the proposed MJP. By shuffling the input data, MJP increases data diversity, which reduces the risk of the model overfitting to specific patterns.

\noindent\textbf{Gradient Leakage.}
Our method disrupts spatial information using jigsaw puzzle input data, which effectively prevent the reconstruction of images and sentences. Both visualization and quantitative comparisons confirm that this approach successfully mitigates the gradient leakage problem. In addition, as the model's generalization improves across different datasets, as shown in Section~\ref{sec:Preserving User Privacy}, the risk of gradient leakage is further minimized. 

\noindent\textbf{\textcolor{black}{Limitations.}}
\textcolor{black}{The application of MJP to domains requiring precise spatial or sequential alignment remains challenging, such as keypoint localization in vision or autoregressive text generation. In these scenarios, shuffling patches or tokens can disrupt the critical structural dependencies that the tasks fundamentally rely on. Therefore, the application of MJP requires careful consideration of the specific task requirements.}

\subsection{Conclusion}
In this paper, we introduce a unified Masked Jigsaw Puzzle (MJP) framework designed to enhance both the performance and the privacy protection of Transformer-based models in vision and text tasks. By providing the model with transformed input data, MJP disrupts the spatial information captured by position embeddings (PEs). This disruption mitigates the risk of gradient inversion attacks without sacrificing model performance on the large-scale datasets. Extensive experiments validate MJP's effectiveness across different domains, which demonstrates its potential as a versatile solution for privacy preservation in federated learning. Future work will focus on further refining the framework and exploring its application in more complex scenarios, such as the multi-modal learning tasks.